%% file: main.tex
\documentclass[10pt,twocolumn,letterpaper]{article}

 \usepackage{cvpr}              

\input{preamble}

\definecolor{cvprblue}{rgb}{0.21,0.49,0.74}
\usepackage[pagebackref,breaklinks,colorlinks,citecolor=cvprblue]{hyperref}

\newcommand{\suseo}[1]{\textcolor{magenta}{#1}}
\newcommand{\jkkim}[1]{\textcolor{brown}{#1}}

\newcommand{\bhr}[1]{\textcolor{red}{#1}}
\newcommand{\bhb}[1]{\textcolor{blue}{#1}}

\usepackage{bbm}
\usepackage{algorithmic}
\usepackage{algorithm}
\usepackage{multirow}
\usepackage{diagbox}

\newtheorem{definition}{Definition}
\newtheorem{proposition}{Proposition}
\newtheorem{assumption}{Assumption}
\usepackage{amsmath}        
\DeclareMathOperator*{\argmin}{arg\,min}

\newtheorem{Proof Sketch}[theorem]{Proof Sketch}

\def\paperID{9311} 
\def\confName{CVPR}
\def\confYear{2024}

\title{Relaxed Contrastive Learning for Federated Learning}

\author{
Seonguk Seo\footnotemark[1]~~$^1$ \qquad  Jinkyu Kim\footnotemark[1]~~$^1$  \qquad Geeho Kim\footnotemark[1]\thanks{indicates equal contribution.}~~$^1$  \qquad Bohyung Han$^{1,2}$ \\
$^1$ECE \& $^{2}$IPAI, Seoul National University\\
 {\tt\small \{seonguk, jinkyu, snow1234, bhhan\}@snu.ac.kr}
}

\begin{document}
\maketitle

\input{sections/abstract.tex}
\input{sections/intro.tex}

\input{sections/related.tex}
\input{sections/prelim.tex}

\input{sections/method.tex}

\input{sections/exp.tex}
\input{sections/conclusion.tex}

\bibliography{cvpr}
\bibliographystyle{cvpr}

\input{sections/supple}

\end{document}

%% file: preamble.tex
\usepackage[dvipsnames]{xcolor}


%% file: sections/abstract.tex

\begin{abstract}

We propose a novel contrastive learning framework to effectively address the challenges of data heterogeneity in federated learning.
We first analyze the inconsistency of gradient updates across clients during local training and establish its dependence on the distribution of feature representations, leading to the derivation of the supervised contrastive learning (SCL) objective to mitigate local deviations.
In addition, we show that a na\"ive integration of SCL into federated learning incurs representation collapse, resulting in slow convergence and limited performance gains. 
To address this issue, we introduce a relaxed contrastive learning loss that imposes a divergence penalty on excessively similar sample pairs within each class.
This strategy prevents collapsed representations and enhances feature transferability, facilitating collaborative training and leading to significant performance improvements.
Our framework outperforms all existing federated learning approaches by significant margins on the standard benchmarks, as demonstrated by extensive experimental results.
The source code is available at our project page\footnote{\url{https://github.com/skynbe/FedRCL}}.

\end{abstract}

%% file: sections/intro.tex

\section{Introduction}
\label{sec:intro}


Federated learning (FL) trains a shared model through the collaboration of distributed clients while safeguarding the privacy of local data by restricting their sharing and transfer.
The primary challenge in this learning framework arises from the data heterogeneity across clients and the class imbalance in local data.
These problems eventually lead to severe misalignments of the local optima of the client models, hindering the search for better global optima of the aggregated model and slowing down convergence. 

To tackle these challenges, most existing approaches focus on minimizing the discrepancy between the global and local models by incorporating regularization techniques on either model parameters~\citep{li2020federated, karimireddy2020scaffold, al2020federated, zhang2020fedpd, acar2021federated} or feature representations~\citep{li2021model, mu2023fedproc, lee2022preservation, yao2021local, kim2022multi}.
However, aligning the local models with the global model entails a trade-off as the global model is not necessarily optimal.
%
%
Recently, there have been several attempts to analyze the inconsistent local training in a principled way~\cite{zhang2022Federated, shi2022towards}.
For example, Zhang~\etal~\cite{zhang2022Federated} investigate the label distribution skewness from a statistical perspective, and introduce a deviation bound for analyzing the inconsistency of gradient updates in local training.
%

We reformulate the deviation bound proposed in~\citep{zhang2022Federated} and establish its dependence on the distribution of feature representations.
Subsequently, we derive that incorporating the supervised contrastive learning (SCL) objective enhances this bound, resulting in consistent local updates across heterogeneous clients.
In other words, we show that employing SCL improves the convergence of federated learning by alleviating the variations of local models.

\begin{figure}[t]
\centering
\hspace{-0.3cm}
    \begin{minipage}[t]{0.5\linewidth}
        \centering
        \includegraphics[width=\linewidth]{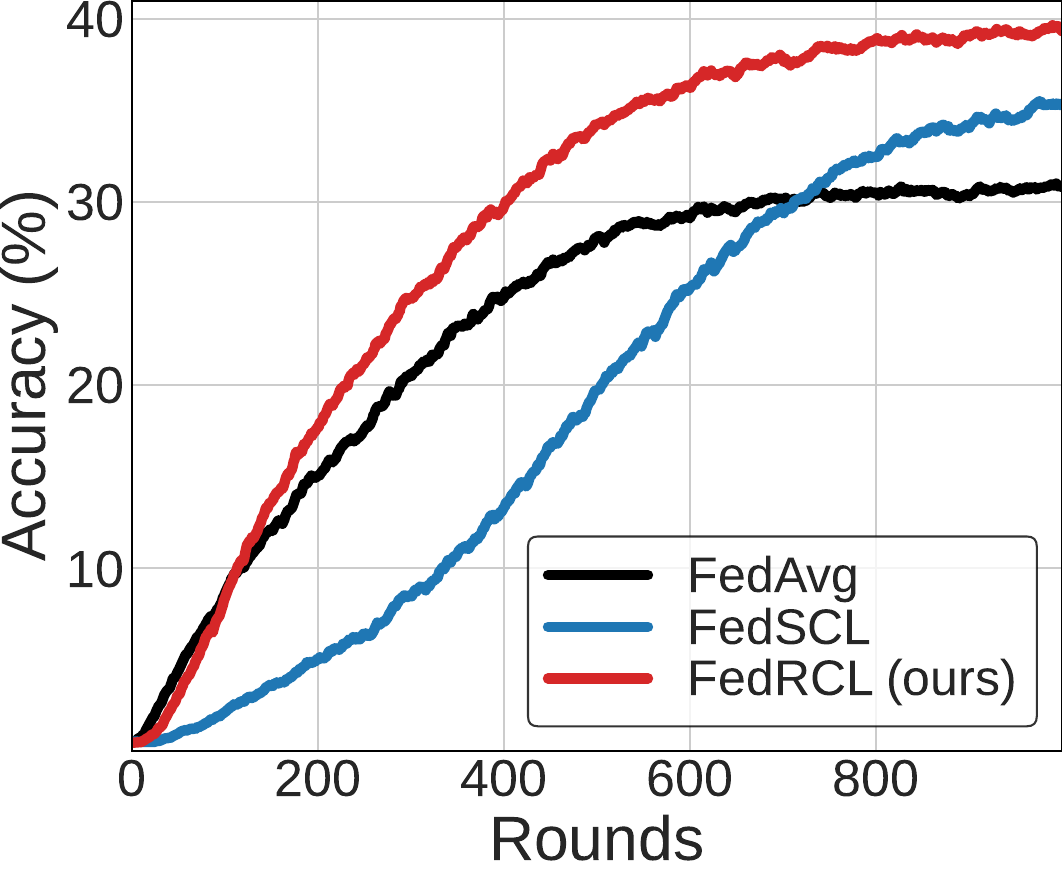}
        \subcaption{Tiny-ImageNet}
        \label{fig:teaser_acc_tiny}
    \end{minipage}
    \hspace{0.1cm}
        \begin{minipage}[t]{0.47\linewidth}
        \centering
        \includegraphics[width=\linewidth]{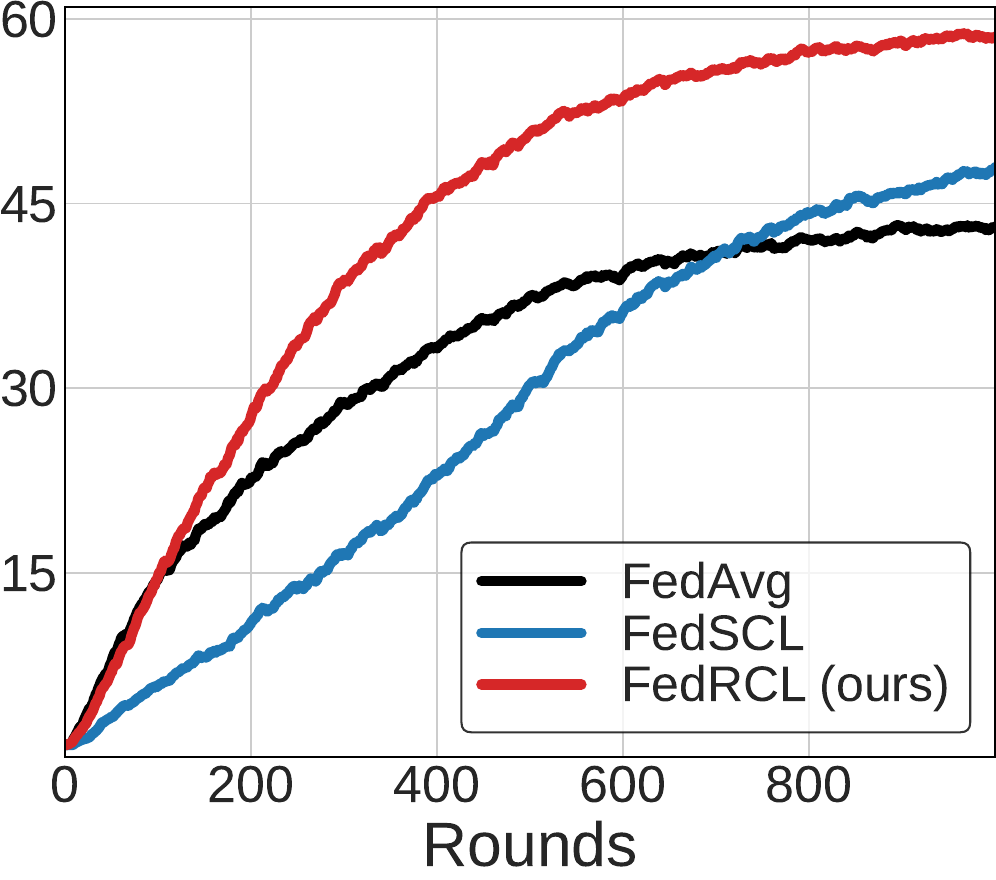}
        \subcaption{CIFAR-100}
        \label{fig:teaser_acc_cifar10}
    \end{minipage}
    \caption{
    Performance curves of our framework, dubbed as FedRCL, in comparison to other baselines on the Tiny-ImageNet and CIFAR-100 with non-\textit{i.i.d.} setting ($\alpha=0.1$).
    FedSCL incorporates the supervised contrastive learning objective into FedAvg, but it suffers from slow convergence and restrains performance enhancement.
Our framework significantly improves both convergence speed and accuracy.
    }
    \label{fig:teaser}
\end{figure}

Although SCL is helpful for the optimization in federated learning, the empirical results show that a na\"ive integration of SCL suffers from slow convergence and limited performance gains as illustrated in Figure~\ref{fig:teaser}.
Due to the limited and imbalanced training data in a local client, the intra-class attraction force in SCL hampers feature diversity and consequently weakens the transferability of neural networks to diverse tasks. 
Considering the consolidation principle of federated learning through the aggregation of heterogeneous local models, the lack of transferability impedes the collaborative training process.

To tackle this issue and enhance the transferability of models, we present a novel contrastive learning strategy for federated learning.
Our approach imposes the penalty on the sample pairs within the same class that may exhibit excessively high similarity otherwise.
Such a simple adaptive repulsion strategy effectively prevents the intra- and inter-class collapse of representations, enhancing the transferability across heterogeneous clients and leading to the discovery of better global optima.
Furthermore, we expand the proposed approach to cover all intermediate levels of representations, promoting consistent local updates even further.
The proposed approach demonstrates remarkable performance improvements in all datasets and settings consistently, surpassing existing baselines by significant margins.
We present the effectiveness and robustness of the proposed method by thorough empirical analysis.
Our main contributions are summarized as follows.

\begin{itemize} 


    \item By reformulating the deviation bound of local gradient update, we theoretically analyze that supervised contrastive learning mitigates inconsistent local updates across heterogeneous clients. \vspace{1mm}

    \item We discover the feature collapse phenomenon caused by the standard SCL in federated learning, resulting in slow convergence and limited performance improvement. \vspace{1mm}

    \item We propose a relaxed supervised contrastive loss, which adaptively imposes the divergence penalty on pairs of examples in the same class and prevents their representations from being learned to be indistinguishable. \vspace{1mm}

    \item We demonstrate that our approach significantly outperforms existing federated learning algorithms on the standard benchmarks under various settings. \vspace{1mm}

    




\end{itemize}

The rest of the paper is organized as follows.
We review the prior works in Section~\ref{sec:related} and discuss the preliminaries in Section~\ref{sec:prelim}.
Section~\ref{sec:method} presents the proposed approach in the context of federated learning and Section~\ref{sec:exp} validates its effectiveness empirically.
Finally, we conclude our paper in Section~\ref{sec:conclusion}.

%% file: sections/related.tex

\section{Related Works}
\label{sec:related}
This section first overviews the existing FL algorithms, and discusses how contrastive learning has been explored in the context of FL.

\subsection{Federated learning}
\label{sub:federated}
McMahan~\etal~\cite{mcmahan2017communication} propose a pioneer FL framework, FedAvg, which aggregates model updates from distributed clients to improve a global model without requiring the exchange of local data.
However, it suffers from slow convergence and poor performance due to the heterogeneous nature of client data in practical scenarios~\citep{zhao2018federated}. 
To address the issue of heterogeneity in FL, numerous approaches have been proposed in two distinct directions, local training and global aggregation.

The major approaches in local training are imposing regularization constraints on model parameters or feature representations.
Specifically, they incorporate proximal terms~\citep{li2020federated}, introduce control variates~\citep{karimireddy2020scaffold, li2019feddane}, or leverage primal-dual analysis~\citep{zhang2020fedpd, acar2021federated} to regularize model parameters, while adopting knowledge distillation~\citep{lee2022preservation, yao2021local, kim2022multi}, metric learning~\citep{zhuang2022divergence, li2021model, mu2023fedproc}, logit calibration~\citep{zhang2022Federated}, feature decorrelation~\citep{shi2022towards}, or data augmentation~\citep{yoon2021fedmix, xu2022acceleration} for effective representation learning.
Our framework also belongs to representation learning, where it particularly focuses on gradient deviations in local training and transferability of trained models across heterogeneous local clients.

Besides the local training methods, server-side optimization techniques have been explored to expedite convergence using momentum~\citep{hsu2019measuring, reddi2021adaptive, kim2022communicationefficient} or decrease the communication cost by quantization~\citep{reisizadeh2020fedpaq, wang2022communication, haddadpour2021federated, nam2022fedpara}.
These server-side works are orthogonal to our client-side approach and are easily combined with the proposed algorithm.

\subsection{Contrastive learning in FL}
\label{sub:contrastive}

Recent works have explored the integration of contrastive learning techniques~\citep{InfoNCE, chen2020simple, he2020momentum} into federated learning to prevent local client drift and assist local training.
FedEMA~\cite{zhuang2022divergence} adopts self-supervised contrastive learning to deal with unlabeled data collected from edge devices.
MOON~\citep{li2021model} introduces a model-contrastive loss, which aims to align the current local model with the global model, while pushing the current model away from the local model of the previous round.
FedProc~\citep{mu2023fedproc} employs a contrastive loss to align local features with the global prototypes to reduce the representation gap, where the global class prototypes are distributed from the server.
FedBR~\citep{guo2023fedbr} conducts contrastive learning to align local and global feature spaces using local data and globally shared proxy data to reduce bias in local training.
In contrast to prior approaches, our framework does not require additional communication overhead for contrastive learning and does not rely on global models or prototypes to mitigate the deviations in local training.

%% file: sections/prelim.tex
\section{Preliminaries}
\label{sec:prelim}
Before discussing the proposed approach, we briefly describe the main idea and formulation of federated learning and supervised contrastive learning.

\subsection{Problem setup}
\label{sec:setup}


Suppose that there are $N$ clients, $\{C_1, ..., C_N\} = \mathcal{C}$. 
Each client $C_i$ has a dataset $\mathcal{D}_i$, which comprises a set of pairs of an example and its class label.
The goal of federated learning is to optimize a global model parametrized by $\theta = [\phi;\psi]$, corresponding to a feature extractor, $\phi$, and a classifier, $\psi$, that minimizes the average losses over all clients as
\begin{align}
    \underset{\theta}{\arg\min} ~ \frac{1}{N}\sum_{i=1}^N \mathcal{L}_i(\theta),
\end{align}
where $\mathcal{L}_i(\theta) = \mathbb{E}_{(\mathbf{x},y)\sim \mathcal{D}_i} \left[ \ell(\mathbf{x}, y;\theta) \right]$ is the empirical loss in $C_i$, given by the expected loss over all samples in $\mathcal{D}_i$.
Note that data distributions in individual clients may be heterogeneous, and privacy concerns strictly prohibit transfering training data across clients.
We employ FedAvg~\citep{mcmahan2017communication} as a baseline algorithm.
In the $t^{\text{th}}$ communication round, a central server sends a global model $\theta^{t-1}$ to the active client set $\mathcal{C}_t \subseteq \mathcal{C}$. 
Each client $C_i \in \mathcal{C}_t$ initializes its parameter $\theta^t_{i,0}$ to $\theta^{t-1}$, and performs $K$ iterations for optimization using its local data.
The server collects the resulting local models $\theta^t_{i,K}$ and computes the global model $\theta^t$ for the next round of training by simply averaging the local model parameters.
This training process is repeated until the global model $\theta^t$ converges.

\subsection{Supervised contrastive learning}
Supervised contrastive learning (SCL)~\cite{khosla2020supervised} is a variant of self-supervised contrastive learning~\cite{chen2020simple, InfoNCE}, where, given the $i^\text{th}$ example and its ground-truth label denoted by $(\mathbf{x}_i, y_i)$, the supervised contrastive loss $\mathcal{L}_\text{SCL}$ is defined as
\begin{align}
\hspace{-3mm} \mathcal{L}_\text{SCL} (\mathbf{x}_i, y_i) =  - \hspace{-1mm} \sum_{\substack{j \neq i, \\ y_j=y_i}} \log \frac{ 
\exp \left( { \langle \phi}(\mathbf{x}_i), \phi(\mathbf{x}_j)\rangle /{\tau} \right)}{\sum\limits_{k \neq i} \exp\left( {\langle \phi(\mathbf{x}_i), \phi(\mathbf{x}_k)\rangle}/{\tau} \right)}, \hspace{-1mm}
\label{eq:scl}
\end{align}
where $\phi(\cdot)$ denotes the feature representation of an input example, $\langle \cdot, \cdot \rangle$ indicates the cosine similarity function, and $\tau$ is a temperature.
To boost its effectiveness, hard example mining is usually adopted to construct both positive and negative pairs.
Eq.~(\ref{eq:scl}) is also expressed as follows:
\begin{align}
\begin{split}
\mathcal{L}_\text{SCL} (\mathbf{x}_i, y_i) = &  \hspace{-1em} {\sum\limits_{y_j=y_i, j\neq i} \hspace{-1em}
 \Big\{- \left( { \langle \phi}(\mathbf{x}_i), \phi(\mathbf{x}_j)\rangle /{\tau} \right)} \\
 & + \log \Big( { \sum\limits_{k \neq i} \exp\left( {\langle \phi(\mathbf{x}_i), \phi(\mathbf{x}_k)\rangle}/{\tau} \right)} \Big)\Big\} .
\end{split}
\label{eq:scl_break}
\end{align}
This loss function encourages feature representations from the same class to be similar while pushing features from different classes apart.

%% file: sections/method.tex

\section{Relaxed Supervised Contrastive Learning}
\label{sec:method}

\begin{figure*}
\centering
    \begin{minipage}[t]{0.305\linewidth}
        \centering
        \includegraphics[width=\linewidth]{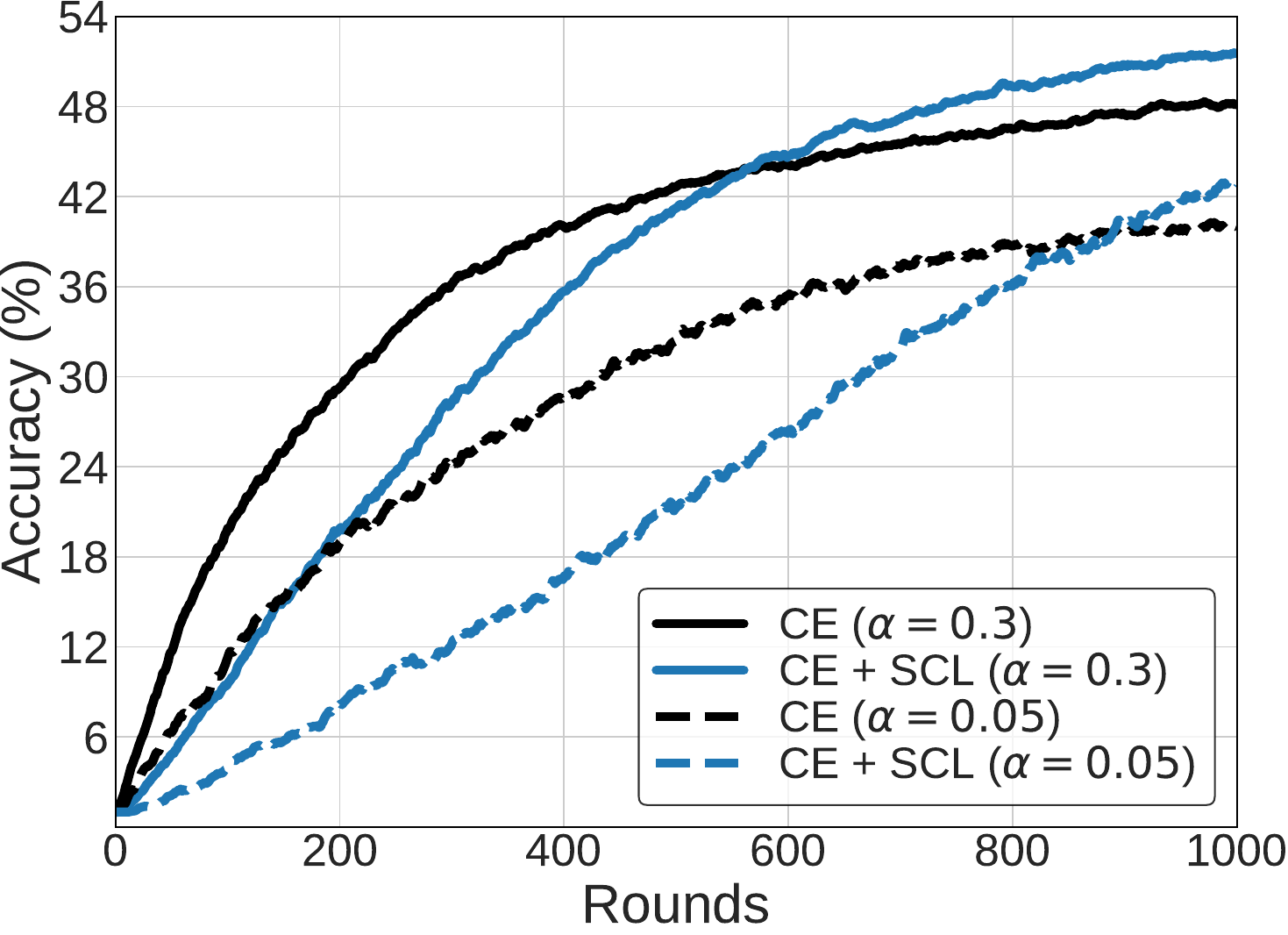}
        \subcaption{Accuracy}
        \label{fig:acc_scl_cifar}
    \end{minipage}
    \hspace{0.26cm}
    \begin{minipage}[t]{0.315\linewidth}
        \centering
        \includegraphics[width=\linewidth]{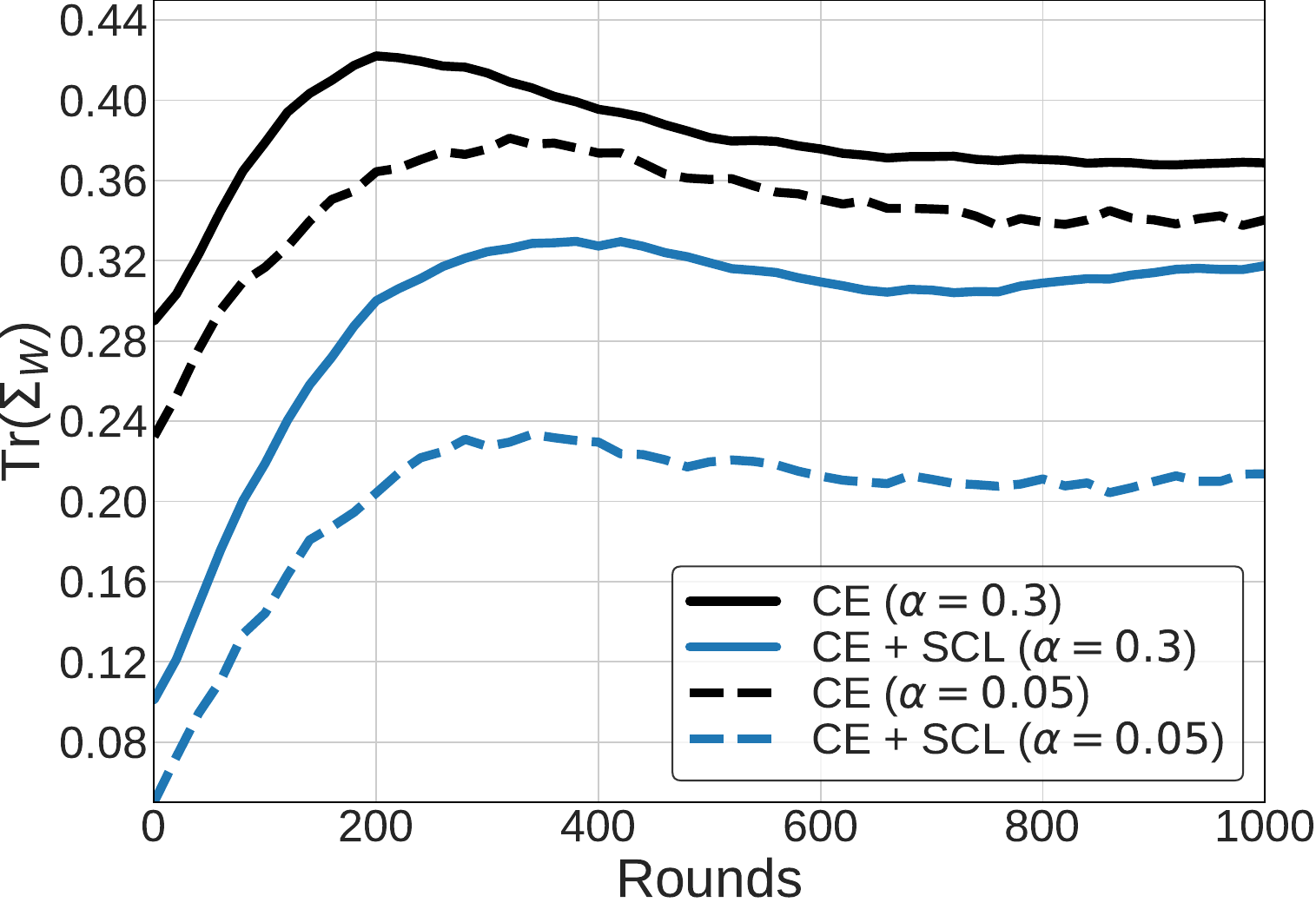}
        \subcaption{Within-class variance}
        \label{fig:within_scl_cifar}
    \end{minipage}
    \hspace{0.22cm}
    \begin{minipage}[t]{0.315\linewidth}
        \centering
        \includegraphics[width=\linewidth]{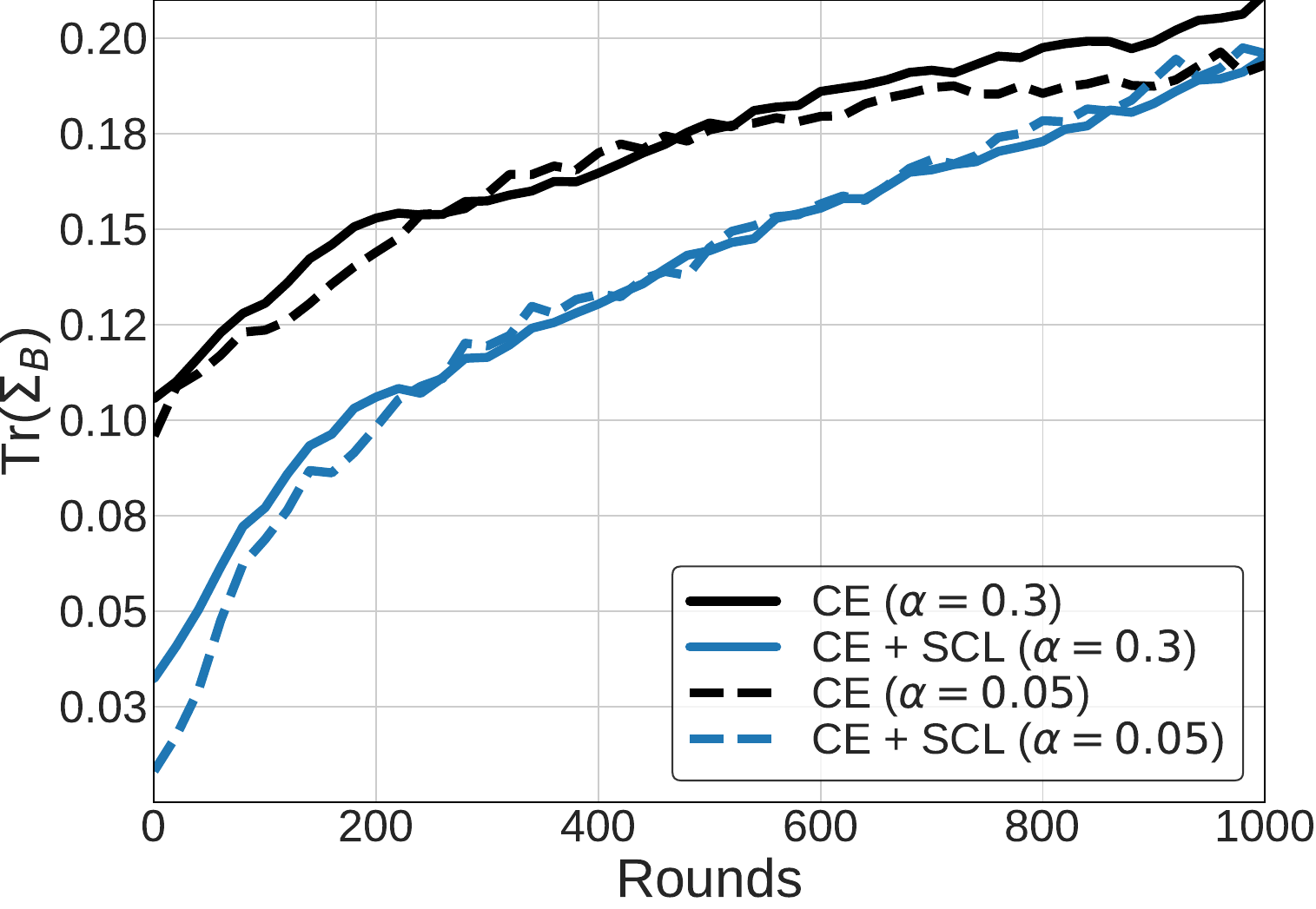}
        \subcaption{Between-class variance}
        \label{fig:between_scl_cifar}
        \vspace{0.3cm}
    \end{minipage}

    \hspace{-0.2cm}
    \begin{minipage}[t]{0.305\linewidth}
        \centering
        \includegraphics[width=\linewidth]{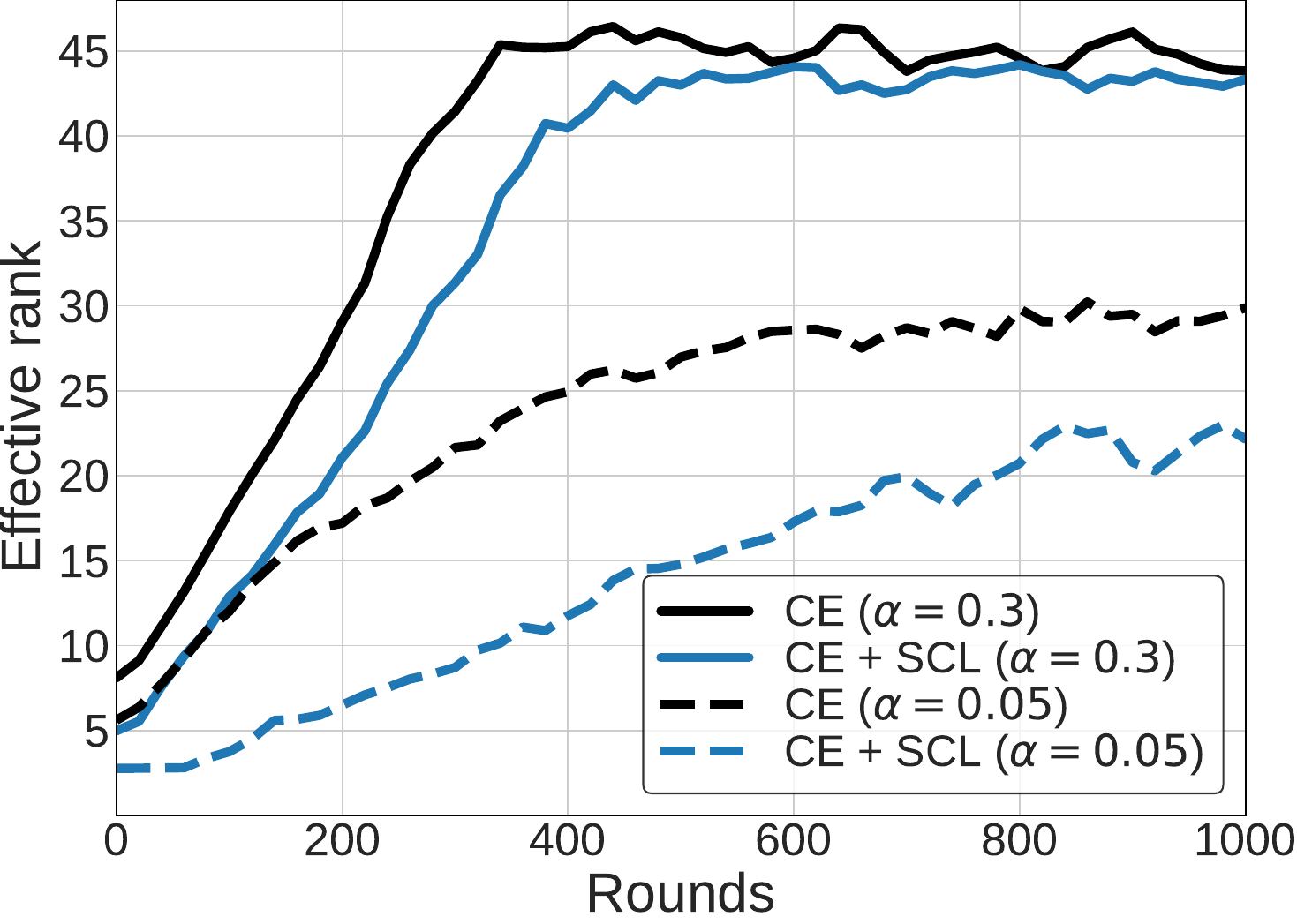}
        \subcaption{Effective rank}
        \label{fig:rank_scl_cifar}
    \end{minipage}
    \hspace{0.42cm}
    \begin{minipage}[t]{0.305\linewidth}
        \centering
        \includegraphics[width=\linewidth]{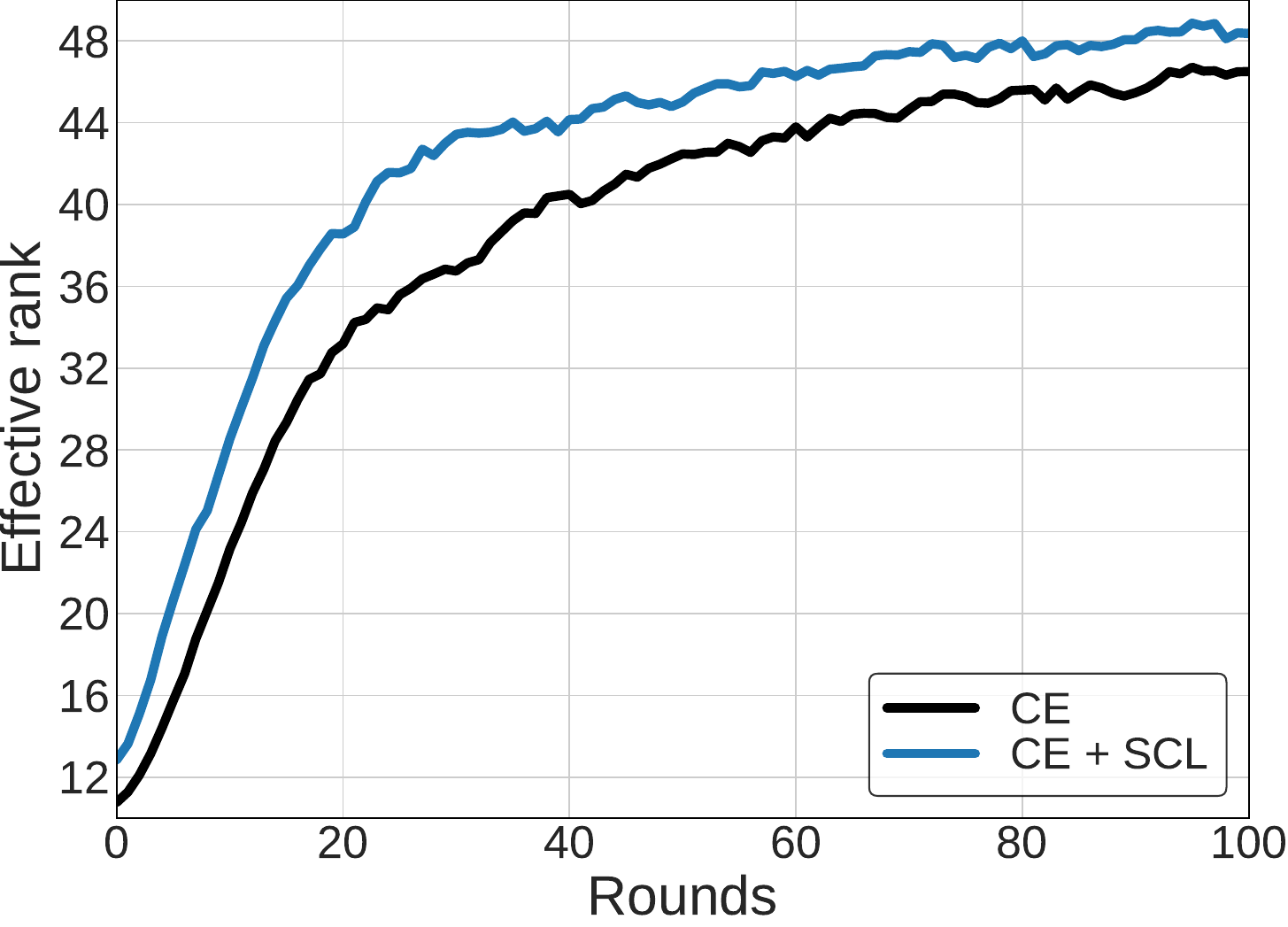}
        \subcaption{Effective rank (centralized setting)}
        \label{fig:rank_scl_cifar_central}
    \end{minipage}
    \hspace{0.28cm}
    \begin{minipage}[t]{0.31\linewidth}
        \centering
        \includegraphics[width=\linewidth]{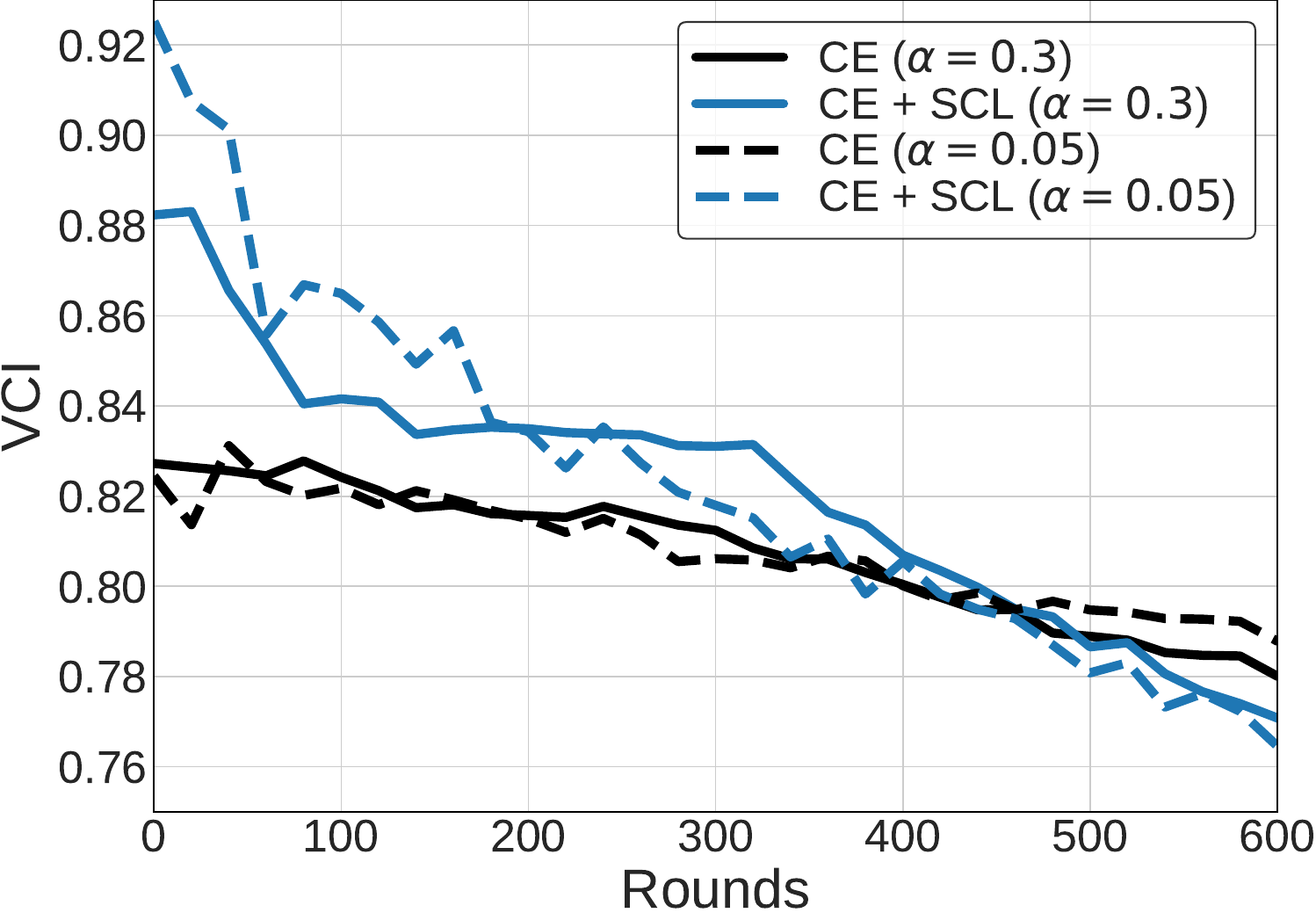}
        \subcaption{Variability collapse index}
        \label{fig:vci_scl_cifar}
    \end{minipage}
    \caption{Effects of employing supervised contrastive loss on the CIFAR-100 under non-\textit{i.i.d.} settings.
    Black and blue lines denote models trained with $\mathcal{L}_\text{CE}$ and $\mathcal{L}_\text{CE}+\mathcal{L}_\text{SCL}$, respectively.
    Dotted and solid lines indicate different data heterogeneity with Dirichlet parameters $\alpha \in \left\{0.05, 0.3\right\}$.
    }
    \label{fig:scl_cifar}
    \vspace{-2mm}
\end{figure*}

%

This section begins by analyzing the local deviations in federated learning with heterogeneous clients, and presents that supervised contrastive learning (SCL) mitigates the deviations (Section~\ref{sec:scl_analysis}).
Then, we identify the challenges in employing SCL in the FL context (Section~\ref{sec:scl_collapse}) and discuss our solution to address the challenges (Section~\ref{sec:rcl} and~\ref{sec:multi_level}).

\subsection{Benefit of SCL for local training}
\label{sec:scl_analysis}

One of the main challenges in federated learning is inconsistent local updates caused by the heterogeneity of local client data. 
Zhang~\etal~\cite{zhang2022Federated} present that existing FL methods based on softmax cross-entropy result in biased local models, and introduce a deviation bound to measure the deviation of the gradient update during the local training.
To analyze this further, we revisit the deviation bound and formulate a sample-wise deviation bound considering all classes, which is formally defined below.

\begin{definition}[\textbf{Sample-wise deviation bound}] \label{samplewisedeviationBound}
Let  $\mathbf{x} \in \mathcal{O}_r$  denote a training example with ground-truth class label $r$. The sample-wise deviation bound is defined as
\begin{align}
\label{eq:deviation_bound}
D(\mathbf{x}) 
& = \frac{\left(1-{P_{r}^{(r)}}\right) {\Phi_{r} |\mathcal{O}_r| S_r(\mathbf{x})}}{\sum\limits_{j \neq r} {P}_{r}^{(j)} {{\Phi_{j}}} |\mathcal{O}_j| S_j(\mathbf{x})},
\end{align}
%
where $P_{z}^{(y)}=\frac{1}{|\mathcal{O}_y|}\sum_{i\in \mathcal{O}_y} p_{z}(\mathbf{x}_{i})$ means the average prediction score for class $z$, estimated with the examples that belong to class $y$, ${\Phi_{y}} = \frac{1}{|\mathcal{O}_y|}\sum_{i \in O_y}{\left\|\phi{(\mathbf{x}_i)}\right\|_{2}}$ is the average feature norm of the examples in class $y$, and $S_y(\mathbf{x}) = \frac{1}{|\mathcal{O}_y|}\sum_{i \in O_y}{\langle \phi{(\mathbf{x})},\phi{(\mathbf{x}_i)} \rangle}$ denotes the average feature similarity with respect to an example $\mathbf{x}$. 
\end{definition}
\begin{proposition} 
If $D(\mathbf{x}) \ll 1$, the local updates of the parameters in classification layer, $\{\Delta \psi_{y}\}_{y \in \mathcal{Y}}$, are prone to deviate from the desirable direction, \ie, $\Delta \psi_{r}\phi(\mathbf{x}) < 0$ and $\exists j \neq r$ such that $\Delta \psi_{j}\phi(\mathbf{x}) > 0$ for $\mathbf{x} \in \mathcal{O}_r$.
\label{prop:local_deviation}
\end{proposition}
%
The proof of Proposition~\ref{prop:local_deviation} is provided in Section~\ref{sec:proof_prop1} of the supplementary document.
Eq.~(\ref{eq:deviation_bound}) indicates that the deviation bound of an example depends on the distribution of feature representations with respect to the example, $S_r(\mathbf{x})$ and $S_j(\mathbf{x})$.
This proposition means that lower values of $D(\mathbf{x})$ incur inconsistent local training.

Proposition~\ref{prop:local_deviation} states that it is possible to prevent the local gradient deviation of each example by increasing $D(\mathbf{x})$.
If $\frac{1}{|\mathcal{Y}|-1} \sum\limits_{j \neq r} S_j(\mathbf{x}) - S_r(\mathbf{x}) \leq 0$, then the lower bound of $D(\mathbf{x})$ in~(\ref{eq:deviation_bound}) becomes $\frac{\left(1-{P_{r}^{(r)}}\right) {\Phi_{r} |\mathcal{O}_r| }}{|\mathcal{Y}|-1} \min\limits_{j \neq r}\left\{\frac{1}{ {P}_{r}^{(j)} {{\Phi_{j}}} |\mathcal{O}_j| }\right\}$.
%
Thus, we formulate the surrogate objective to minimize $\max \Big(0, \frac{1}{|\mathcal{Y}|-1} \sum\limits_{j \neq r} S_j(\mathbf{x}) - S_r(\mathbf{x}) \Big)$, which is highly correlated to the increase of the lower bound.
By using a smooth approximation to the maximum function with the \textit{LogSumExp} operator, we derive its upper bound as follows
\begin{align}
&\max\Big(0, \sum\limits_{j \neq r} \frac{S_j(\mathbf{x})}{|\mathcal{Y}|-1} - S_r(\mathbf{x}) \Big) \nonumber\\
&\leq \log\Big(\exp(0) + \exp \Big( \sum_{j \neq r}\frac{S_j(\mathbf{x})}{|\mathcal{Y}|-1}  - S_r(\mathbf{x}) \Big) \Big) \nonumber \\
&\leq -\frac{1}{|\mathcal{O}_r|-1} \hspace{-0.1cm} \sum_{\mathbf{x}_i \in \mathcal{O}_r \setminus \mathbf{x}} \hspace{-0.1cm}\log\left( \frac{\exp( {\langle \phi{(\mathbf{x})},\phi{(\mathbf{x}_i)} \rangle})}{\sum_{\mathbf{x}_k \neq \mathbf{x}}\exp({\langle \phi{(\mathbf{x})},\phi{(\mathbf{x}_k)} \rangle})} \right).  \nonumber
\end{align}
Please refer to Section~\ref{sec:proof_prop2} for further details.
This derivation demonstrates how the optimization of $\mathcal{L}_\text{SCL}$ contributes to mitigating local gradient deviations.

\begin{figure*}[t!]
\centering
    \begin{minipage}[t]{0.24\linewidth}
        \centering
        \includegraphics[width=\linewidth]{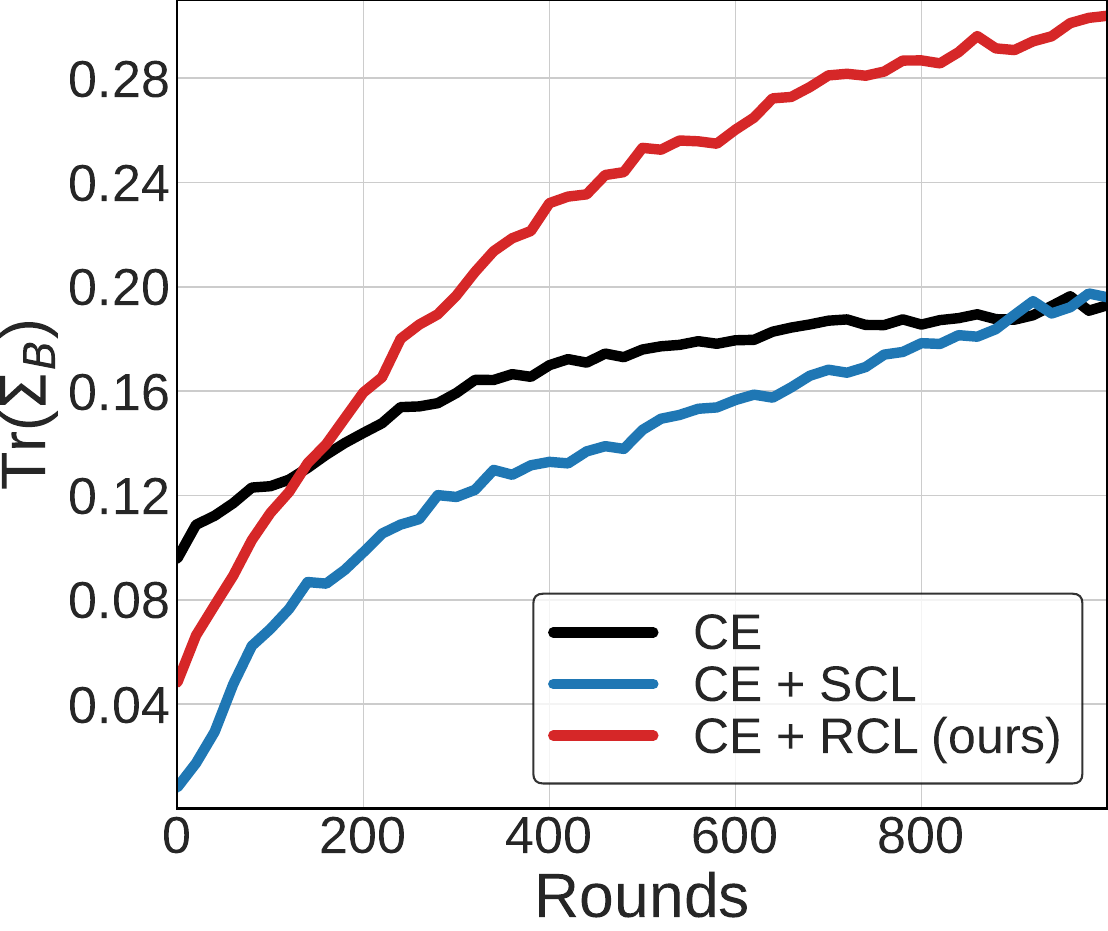}
        \subcaption{Between-class variance}
        \label{fig:between_ucl_cifar}
    \end{minipage}
    \hspace{0.1cm}
    \begin{minipage}[t]{0.23\linewidth}
        \centering
        \includegraphics[width=\linewidth]{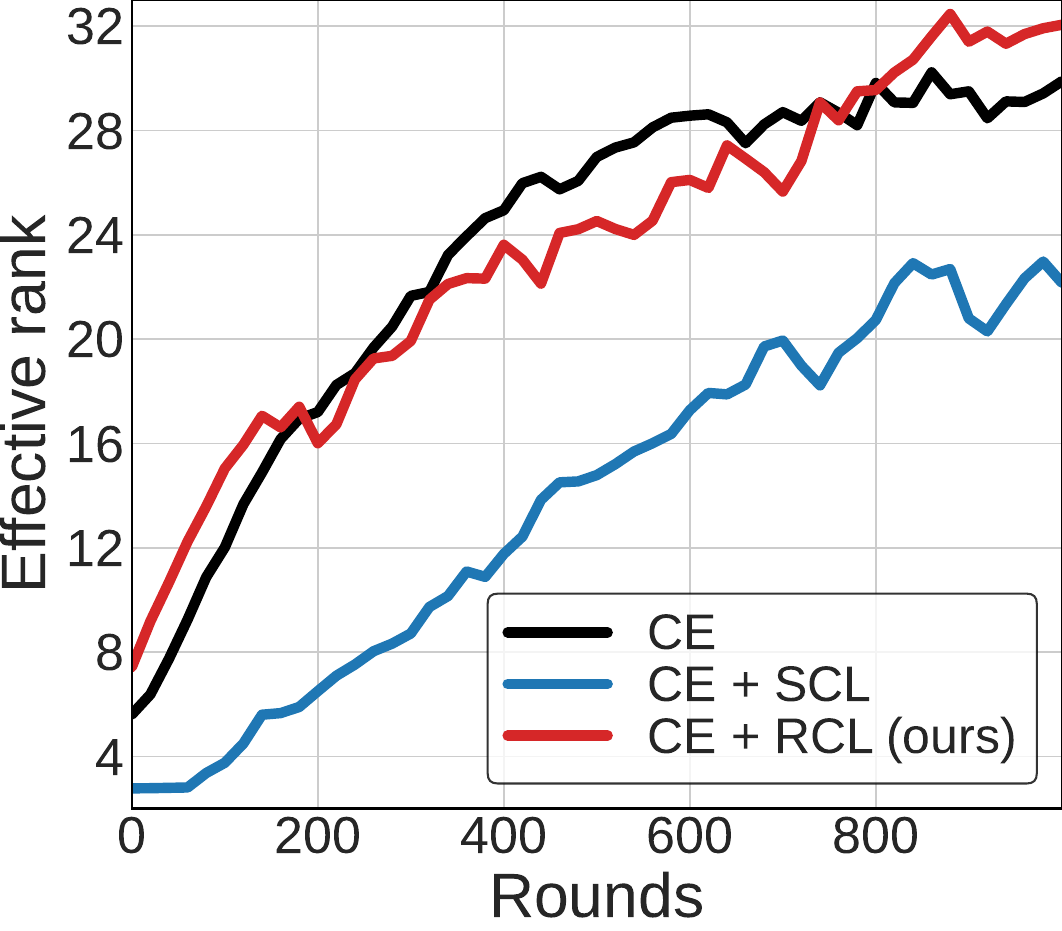}
        \subcaption{Effective rank}
        \label{fig:rank_ucl_cifar}
    \end{minipage}
    \hspace{0.1cm}
    \begin{minipage}[t]{0.24\linewidth}
        \centering
        \includegraphics[width=\linewidth]{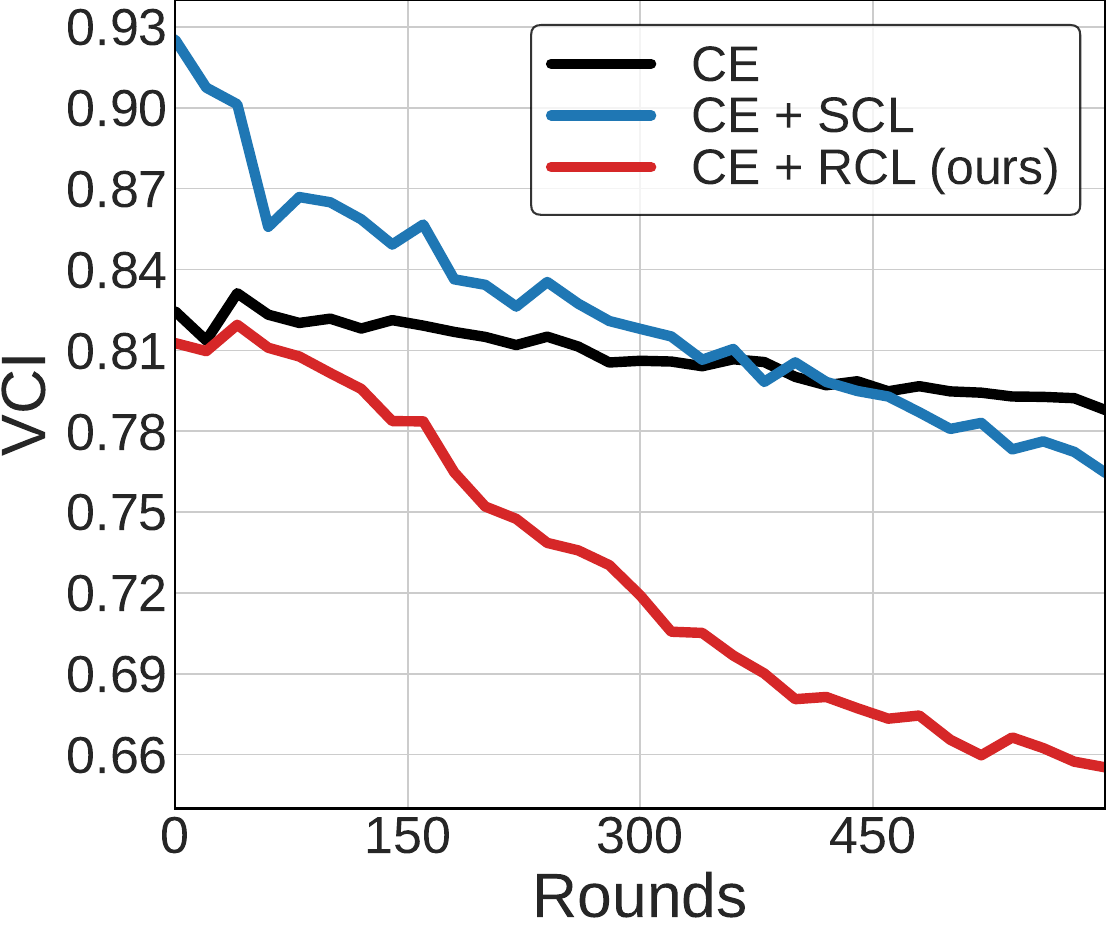}
        \subcaption{Variability collapse index}
        \label{fig:vci_ucl_cifar}
    \end{minipage}
    \hspace{0.1cm}
    \begin{minipage}[t]{0.23\linewidth}
        \centering
        \includegraphics[width=\linewidth]{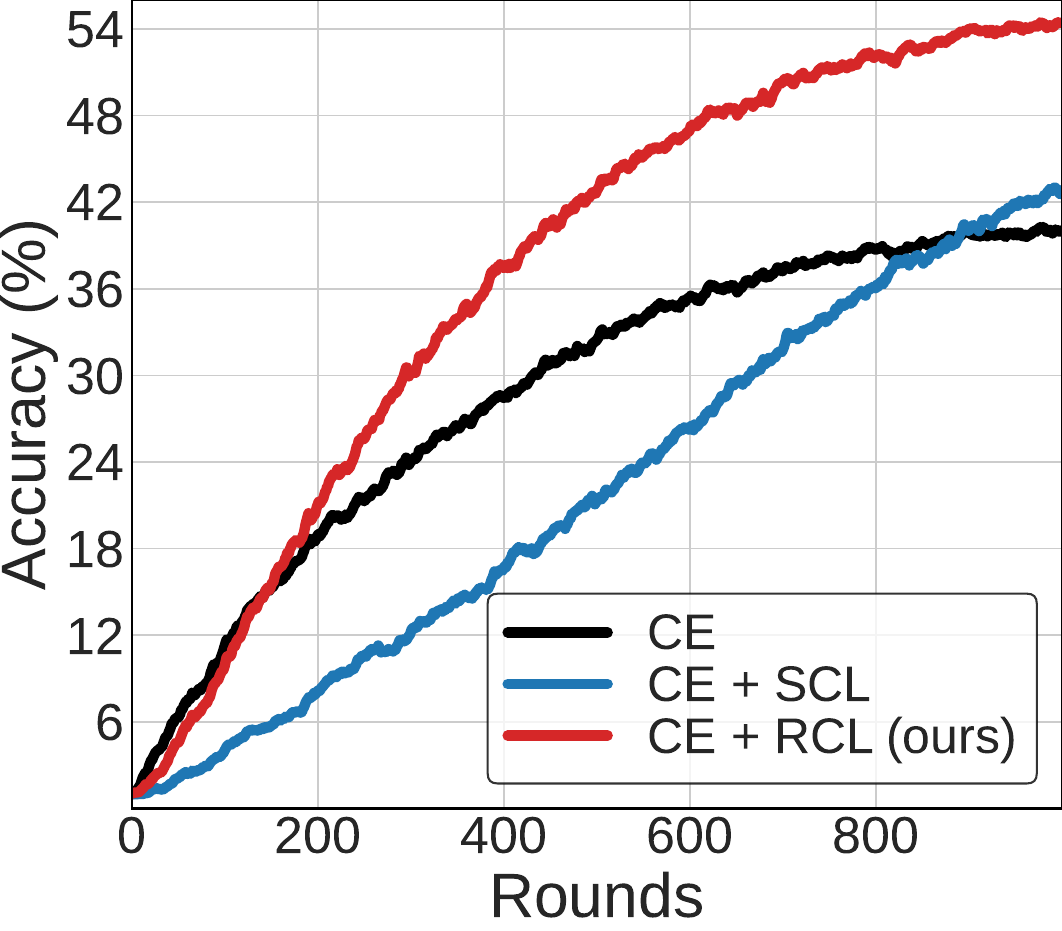}
        \subcaption{Accuracy}
        \label{fig:acc_ucl_cifar}
    \end{minipage}
    
    \caption{Results of our relaxed contrastive learning (RCL) approach on the CIFAR-100 dataset under a non-\textit{i.i.d.} setting ($\alpha=0.05$).
    RCL outperform SCL in all metrics.
    }
    \label{fig:ucl_cifar}
    \vspace{-2mm}
\end{figure*}

\subsection{Representation collapse in FL with SCL}
\label{sec:scl_collapse}

Based on our analysis in Section~\ref{sec:scl_analysis}, we empirically validate the effectiveness of SCL in federated learning under data heterogeneity with Dirichlet parameters $\alpha \in \{0.05, 0.3\}$ in Figure~\ref{fig:acc_scl_cifar}.
We train the ResNet-18 model using the loss function $\mathcal{L} = \mathcal{L}_\text{CE} + \mathcal{L}_\text{SCL}$ at each local client on the CIFAR-100 dataset, using 5\% participation rate out of 100 distributed clients, where $\mathcal{L}_\text{CE}$ represents the cross-entropy loss.
As depicted in the figure, while SCL eventually achieves improved performance by reducing local deviations over the baseline methods, it is accompanied by a noticeable lag in the convergence speed during at early stage of training.
We conjecture that, due to limited and skewed local training data, SCL leads to excessively compact representations of the examples in the same classes, hindering effective knowledge transfer across clients in federated learning.
%

To delve into these phenomena, we first compute the within-class and between-class covariance matrices of the feature embeddings provided by a local model, denoted by $\mathbf{\Sigma}_W$ and $\mathbf{\Sigma}_B$, respectively.
Figure~\ref{fig:within_scl_cifar} and~\ref{fig:between_scl_cifar} plot the trace of the two matrices.
SCL effectively reduces the within-class variance compared to the baseline model only with the cross-entropy loss, due to the attraction term between samples from the same class.
However, it is noteworthy that SCL also yields a lower between-class variance than the baseline, especially at the early stages of training, despite the repulsion term between the examples in different classes.
Since the attraction and repulsion forces interact in contrastive learning, the excessive representation similarity between positive pairs weakens the repulsion force between negative pairs.
In other words, the collapse of intra-class representations negatively affects the separation between inter-class examples, leading to an overall reduction in the diversity of feature representations.
To evaluate this feature collapse quantitatively, we observe the effective rank~\cite{roy2007effective} of the covariance matrix of all feature embeddings given by a local model, which estimates the actual dimensionality of the learned feature manifold of training data.
Formally, the effective rank is defined as follows:

\begin{definition}[\textbf{Effective rank}] \label{def:effective_rank}
Consider a matrix $\mathbf{A} \in \mathbb{R}^{m \times n}$ with its singular values $\{\sigma_1, ..., \sigma_Q\}$, where $Q = \min(m, n)$, and 
let $\mathrm{p}_k = {\sigma_k}/{\sum_{i=1}^{Q}|\sigma_i|}$.
Then, the effective rank of matrix $\mathbf{A}$ is defined as $\exp({H(\mathrm{p}_1, ..., \mathrm{p}_Q)}) = \exp(-\hspace{-0.1cm}\sum_{k=1}^{Q} \mathrm{p}_k \log \mathrm{p}_k)$, where $H(\hspace{-0.02cm}\cdot\hspace{-0.02cm})$ is the Shannon entropy.
\end{definition}

Figure~\ref{fig:rank_scl_cifar} illustrates the impact of SCL on the effective rank in the CIFAR-100 test set.
It supports that SCL diminishes the effective rank when compared to the baseline methods, particularly during the early stage of training, and leads to overall representation collapses.
Interestingly, SCL does not exhibit dimensional collapse in the centralized setting\footnote{We trained a ResNet-18 model with a single client using the whole CIFAR-100 training set.} as in Figure~\ref{fig:rank_scl_cifar_central}, which implies that limited and skewed local training data incurs the problem in SCL.

These collapsed representations exacerbate the transferability of neural networks across heterogeneous tasks and clients.
Previous studies~\citep{cui2022discriminability, sariyildiz2022improving, xu2023quantifying} have emphasized the close relationship between feature diversity and transferability, highlighting that representation collapses of trained models hamper maintaining crucial information beneficial for knowledge transfer to downstream tasks.
To quantitatively analyze this, we employ the variability collapse index~\citep{xu2023quantifying}, $\text{VCI} = 1 - \frac{\text{Tr}[\Sigma_T^\dagger \Sigma_B]}{\text{rank}(\Sigma_B)}$, where $\Sigma_T$ and $\Sigma_B$ denote the total covariance and between-class covariance matrices for a given feature matrix.
It provides a robust measurement of transferability in terms of optimal linear probing loss, where lower values denote better transferability.
As observed in Figure~\ref{fig:vci_scl_cifar}, SCL yields higher VCI values at the early stage, indicating low transferability even in comparison to the baselines.
Given that federated learning can be regarded as a continual fine-tuning process across heterogeneous local tasks, the lack of transferability impedes collaborative training, resulting in slow convergence and limited performance gain.
We will discuss strategies for addressing these challenges in the following subsection.

\begin{algorithm}[t]
   \caption{FedRCL}
   \label{alg:proposed_method}
\begin{algorithmic}[1]
   \STATE {\bfseries Input:} {initial  model $\theta^0$, \# of communication rounds $T$, \\ \quad\quad\quad \# of local iterations $K$,  \# of layers $L$}
   \FOR{$\text{each round}~t = 1, \dots ,T$}
   \STATE Sample a subset of clients $\mathcal{C}_t \subseteq \mathcal{C}$
   \STATE Server sends $\theta^{t-1}$ to all active clients $C_i \in \mathcal{C}_t$
   \FOR{$\text{each}~ C_i \in \mathcal{C}_t,~\textbf{in parallel}$}
   \STATE $\theta_{i,0}^t \leftarrow {\theta^{t-1}}$
   \FOR{$k = {1, \dots ,K}$}
   \FOR {{\bf{each}} $(\mathbf{x},y)$ in a batch}
   \STATE $\mathcal{L}_\text{RCL} \leftarrow \frac{1}{L} \sum_{l=1}^{L} \mathcal{L}_\text{RCL}(\mathbf{x}, y; \phi_l)$
   \STATE $\mathcal{L}(\theta_{i,k-1}^t) \leftarrow \mathcal{L}_{\text{CE}} + \mathcal{L}_\text{RCL}$
   \STATE ${\theta_{i,k}^t \leftarrow \theta_{i,k-1}^t - \eta \nabla \mathcal{L} (\theta_{i,k-1}^t)}$ 
   \ENDFOR
   
   \ENDFOR
   \STATE Client sends  $\theta_{i,K}^t$ back to the server
   \ENDFOR
   \STATE {\bfseries In server:} \\ \quad {$\theta^{t}$ = $\frac{1}{|\mathcal{C}_t|} \sum_{C_i \in \mathcal{C}_t}\theta_{i,K}^{t}$}
   \ENDFOR
\end{algorithmic}
\end{algorithm}


\input{./tables/main_table.tex}


\subsection{Relaxed contrastive loss for FL}
\label{sec:rcl}

To address the representation collapse issue identified in Section~\ref{sec:scl_collapse}, we propose a novel federated learning approach with an advanced contrastive learning strategy, referred to as Federated Relaxed Contrastive Learning (FedRCL).
The proposed algorithm adopts the relaxed contrastive loss $\mathcal{L}_\text{RCL}$, imposing the feature divergence on intra-class samples as 
%
\begin{align}
& \mathcal{L}_{\text{RCL}} (\mathbf{x}_i,y_i;\phi) = \hspace{-1mm} \sum_{\substack{j \neq i, \\ y_j=y_i}} 
\Big\{
- \log \frac{ 
\exp \left( { \langle \phi}(\mathbf{x}_i), \phi(\mathbf{x}_j)\rangle /{\tau} \right)}{\sum\limits_{k \neq i} \exp\left( {\langle \phi(\mathbf{x}_i), \phi(\mathbf{x}_k)\rangle}/{\tau} \right)}  \nonumber \\
& \hspace{3mm} + \beta \cdot
\log \Big( {   \sum_{\substack{{\mathbf{x}_k\in \mathcal{P}(\mathbf{x}_i)}}} \exp\Big( {\langle \phi(\mathbf{x}_i), \phi(\mathbf{x}_k)\rangle}/{\tau} \Big)} + \exp (1/{\tau})  \Big)
\Big\}
\label{eq:rcl}
\end{align}
%
where $\mathcal{P}(\mathbf{x}) = \left\{ {\mathbf{x}'}|y_{{\mathbf{x}'}} = y_{\mathbf{x}}, \langle \phi({\mathbf{x}'}), \phi(\mathbf{x})\rangle > \lambda   \right\}$ represents a set of intra-class samples more similar to the anchor $\mathbf{x}$ than the threshold $\lambda$ and $\beta$ is a hyperparameter for the divergence term.
%
The second term of Eq.~(\ref{eq:rcl}) serves to prevent within-class representation collapses, which also promotes the separation of the examples between different classes. 
This ultimately enhances overall feature diversity and transferability, which is crucial in the context of federated learning with non-\textit{i.i.d.}~settings.
As illustrated in Figure~\ref{fig:ucl_cifar}, FedRCL facilitates inter-class separation, mitigates dimensional collapse, and improves the transferability of trained models, resulting in early convergence and significant performance improvement.

\subsection{Multi-level contrastive training}
\label{sec:multi_level}
Existing contrastive learning approaches~\citep{chen2020simple, InfoNCE, khosla2020supervised} concentrate on aligning the feature representations of the last layer, resulting in predominant model updates in deeper layers while having limited influence on lower-layer parameters.
%
To mitigate this issue, we expand the proposed contrastive learning approach to encompass feature representations in earlier layers.
Let $\phi_l(\mathbf{x})$ denotes the $l^\text{th}$ level feature representation of sample $\mathbf{x}$.
Then, we construct $\mathcal{L}_\text{RCL}$ by aggregating $\frac{1}{L}\sum_{l=1}^{L} \mathcal{L}_\text{RCL}(\mathbf{x}, y; \phi_l)$, where $L$ is the number of layers.
The comprehensive algorithm of our framework is presented in Algorithm~\ref{alg:proposed_method}.


\input{./tables/client_table.tex}


\subsection{Discussion}

FedRCL has something common with existing methods incorporating contrastive loss for local updates, but it has clear differences and advantages over them.
While most existing works~\cite{li2021model,mu2023fedproc,guo2023fedbr} employ contrastive learning to regulate local training towards the global model for consistent local updates, this constraint often leads to suboptimal solutions as the global model is not fully optimized.
Our approach is free from this issue, because FedRCL mitigates local deviations by itself so it does not align with the global model explicitly.
Some algorithms require proxies for contrastive learning such as global prototypes~\cite{mu2023fedproc} or globally shared data~\cite{guo2023fedbr}, which rely on extra communication overhead and full client participation.
Note that transferring such prototypes or data is vulnerable and incurs privacy concerns.
In contrast, FedRCL does not involve any additional overhead and consistently demonstrates strong performance improvement even with an extremely low participation rate.

%% file: tables/main_table.tex

\begin{table*}[t]
\begin{center}
\caption{Results from 5\% participation rate over 100 distributed clients on the CIFAR-10, CIFAR-100, and Tiny-ImageNet for the different levels of Dirichlet parameter ($\alpha$).
Accuracies at the target round are based on the exponential moving average results with parameter $0.9$.}
\vspace{-2mm}
\label{tab:cifar100}
\setlength\tabcolsep{7.9pt}
\hspace{-0.2cm}
\scalebox{0.85}{
\begin{tabular}{clcccccccccc} 
\toprule
\multirow{2}{*}{Dataset} & \multirow{2}{*}{Method}                                  
 & \multicolumn{2}{c}{$\alpha = 0.05$} & \multicolumn{2}{c}{$\alpha = 0.1$} & \multicolumn{2}{c}{$\alpha = 0.3$} & \multicolumn{2}{c}{$\alpha = 0.6$} & \multicolumn{2}{c}{\textit{i.i.d.}}  \\
& & \multicolumn{1}{c}{500R} & \multicolumn{1}{c}{1000R} & \multicolumn{1}{c}{500R} & \multicolumn{1}{c}{1000R}& \multicolumn{1}{c}{500R} & \multicolumn{1}{c}{1000R}   &\multicolumn{1}{c}{500R} & \multicolumn{1}{c}{1000R} &\multicolumn{1}{c}{500R} & \multicolumn{1}{c}{1000R}             \\ 
\midrule
\multirow{10}{*}{CIFAR-10} & FedAvg~\cite{mcmahan2017communication} & 51.47 & 63.42 & 58.80 & 70.82 & 75.63 & 83.18 & 80.93 & 85.52 & 84.67 & 88.19 \\
& FedAvg + FitNet~\cite{romero2014fitnets} & 51.34 & 63.25 & 58.67 & 71.09 & 74.87 & 83.03 & 79.14 & 84.84 & 84.20 & 87.67 \\
& FedProx~\cite{li2020federated} & 48.61 & 59.58 & 56.22 & 68.87 & 70.30 & 80.46 & 76.06 & 83.48 & 84.14 & 87.66 \\
& MOON~\cite{li2021model} & 49.68 & 61.73 & 69.16 & 77.12 & 83.32 & 86.30 & 84.95 & 87.99 & 88.24 & 89.66 \\
& FedMLB~\cite{kim2022multi} & 32.81 & 49.16 & 52.01 & 72.31 & 74.98 & 84.08 & 77.84 & 85.96 & 86.84 & 89.93 \\
& FedLC~\citep{zhang2022Federated} & 54.30 & 65.62 & 62.39 & 72.52 & 78.37 & 84.79 & 81.17 & 86.02 & 84.57 & 88.41 \\
& FedNTD~\cite{lee2022preservation} & 52.33 & 63.36 & 62.23 & 73.54 & 76.05 & 83.78 & 81.20 & 86.46 & 85.98 & 89.44 \\
& FedProc~\cite{mu2023fedproc}  & 25.61 & 47.77 & 33.28 & 62.56 & 63.03 & 80.93 & 69.41 & 84.57 & 78.30 & 87.66 \\
& FedDecorr~\citep{shi2022towards} & 53.04 & 66.62 & 63.74 & 75.35 & 76.62 & 83.40 & 81.39 & 85.28 & 85.41 & 88.16 \\
& \textbf{FedRCL (ours)} & \textbf{64.44} & \textbf{76.74} & \textbf{74.82} & \textbf{82.72} & \textbf{84.01} & \textbf{88.44} & \textbf{86.00} & \textbf{89.45} & \textbf{89.70} & \textbf{91.90} \\
\midrule
\multirow{10}{*}{CIFAR-100} & FedAvg~\cite{mcmahan2017communication} & 31.20 & 39.86 & 36.65 & 43.04 & 41.70 & 47.47 & 43.23 & 49.29 & 43.52 & 48.12 \\
& FedAvg + FitNet~\cite{romero2014fitnets} & 31.09 & 38.35 & 36.48 & 43.25 & 42.96 & 48.59 & 44.20 & 49.82 & 44.61 & 49.33 \\
& FedProx~\cite{li2020federated} & 30.27 & 39.44 & 35.78 & 43.11 & 42.24 & 48.19 & 43.21 & 48.48 & 45.20 & 49.37 \\
& MOON~\cite{li2021model} & 34.28 & 40.64 & 42.91 & 50.31 & 53.15 & 58.37 & 55.76 & 61.42 & 58.50 & 64.73 \\
& FedMLB~\cite{kim2022multi} & 30.89 & 43.89 & 38.64 & 48.94 & 47.39 & 54.58 & 49.36 & 56.70 & 50.12 & 56.40 \\
& FedLC~\citep{zhang2022Federated} & 34.24 & 40.84 & 39.80 & 44.40 & 42.74 & 47.23 & 44.24 & 48.89 & 44.06 & 47.63 \\
& FedNTD~\cite{lee2022preservation} & 33.10 & 41.75 & 35.84 & 42.86 & 43.22 & 49.29 & 44.26 & 50.32 & 44.93 & 50.15 \\
& FedProc~\cite{mu2023fedproc}  & 18.41 & 38.56 & 25.19 & 43.73 & 32.66 & 49.68 & 36.09 & 49.89 & 40.76 & 52.94 \\
& FedDecorr~\citep{shi2022towards} & 33.31 & 41.73 & 38.88 & 43.89 & 43.52 & 49.17 & 44.01 & 49.08 & 45.46 & 49.30 \\
& \textbf{FedRCL (ours)} & \textbf{43.71} & \textbf{54.63} & \textbf{49.82} & \textbf{58.23} & \textbf{57.89} & \textbf{63.46} & \textbf{58.71} & \textbf{64.06} & \textbf{60.25} & \textbf{64.81} \\
\midrule
\multirow{10}{*}{Tiny-ImageNet} & FedAvg~\cite{mcmahan2017communication} & 22.49 & 25.90 & 26.62 & 29.71 & 31.80 & 33.58 & 33.91 & 35.01 & 35.62 & 37.02 \\
& FedAvg + FitNet~\cite{romero2014fitnets} & 22.82 & 26.95 & 27.37 & 30.51 & 32.96 & 33.95 & 33.46 & 34.70 & 35.79 & 37.31 \\
& FedProx~\cite{li2020federated} & 22.91 & 27.02 & 27.31 & 30.93 & 32.35 & 34.34 & 34.33 & 35.53 & 35.94 & 36.11 \\
& MOON~\cite{li2021model} & 23.30 & 26.34 & 30.31 & 32.03 & 36.97 & 39.32 & 38.98 & 42.07 & 41.88 & 45.62 \\
& FedMLB~\cite{kim2022multi} & 19.31 & 26.88 & 29.31 & 34.41 & 37.20 & 40.16 & 39.34 & 42.15 & 40.69 & 42.98 \\
& FedLC~\citep{zhang2022Federated} & 26.30 & 28.28 & 30.63 & 32.25 & 35.03 & 35.95 & 35.38 & 36.48 & 36.57 & 37.75 \\
& FedNTD~\cite{lee2022preservation} & 22.83 & 28.96 & 28.86 & 33.74 & 33.91 & 37.33 & 36.47 & 39.43 & 37.77 & 40.85 \\
& FedProc~\cite{mu2023fedproc}  & 10.74 & 22.74 & 14.02 & 27.43 & 16.62 & 32.43 & 19.64 & 32.60 & 21.59 & 35.43 \\
& FedDecorr~\citep{shi2022towards} & 22.55 & 26.18 & 28.15 & 30.74 & 33.40 & 34.86 & 33.31 & 34.90 & 35.02 & 35.82 \\
& \textbf{FedRCL (ours)} & \textbf{27.21} & \textbf{34.60} & \textbf{34.30} & \textbf{39.36} & \textbf{40.25} & \textbf{44.95} & \textbf{43.20} & \textbf{46.70} & \textbf{45.01} & \textbf{47.25} \\
\bottomrule
\end{tabular}}
\end{center}
\vspace{-3mm}
\end{table*}

%% file: tables/client_table.tex

\begin{table*}[t!]
\begin{center}
\caption{Results from 2\% participation rate over 100 and 500 clients on three benchmarks.
The Dirichlet parameter is commonly set to $0.3$.}
\label{tab:large_scale}
\vspace{-2mm}
\setlength\tabcolsep{7.7pt}
\scalebox{0.85}{\begin{tabular}{lcccc|cccc|cccc} 
\toprule
& \multicolumn{4}{c}{CIFAR-10}  & \multicolumn{4}{c}{CIFAR-100}  & \multicolumn{4}{c}{Tiny-ImageNet} \\ 
 & \multicolumn{2}{c}{100 clients} & \multicolumn{2}{c|}{500 clients} & \multicolumn{2}{c}{100 clients} & \multicolumn{2}{c|}{500 clients} & \multicolumn{2}{c}{100 clients} & \multicolumn{2}{c}{500 clients}  \\
Method & {500R} & {1000R} & {500R} & {1000R} & {500R} & {1000R} & {500R} & {1000R} & {500R} & {1000R} & {500R} & {1000R}  \\
\midrule
FedAvg~\cite{mcmahan2017communication} & 65.92 & 78.13 & 59.88 & 72.12 & 38.19 & 44.62 & 29.01 & 37.86 & 28.63 & 34.62 & 21.00 & 27.37 \\
FedAvg + FitNet~\cite{romero2014fitnets} & 66.88 & 79.22 & 57.29 & 70.94 & 36.89 & 46.69 & 28.52 & 36.41 & 27.80 & 34.88 & 20.17 & 27.10 \\
FedProx~\cite{li2020federated} & 65.78 & 75.82 & 60.23 & 72.78 & 36.69 & 45.16 & 28.44 & 35.45 & 27.45 & 32.91 & 22.34 & 29.04 \\
MOON~\cite{li2021model} & 71.52 & 75.42 & 69.15 & 78.06 & 39.91 & 46.51 & 33.51 & 42.41 & 27.26 & 32.25 & 26.69 & 31.81 \\
FedMLB~\cite{kim2022multi} & 65.85 & 79.45 & 58.68 & 71.38 & 40.90 & 53.34 & 32.03 & 42.61 & 31.17 & 38.09 & 28.39 & 33.67 \\
FedLC~\citep{zhang2022Federated} & 72.90 & 80.90 & 60.16 & 71.39 & 39.70 & 42.10 & 29.58 & 36.78 & 30.94 & 35.59 & 22.14 & 26.83 \\
FedNTD~\cite{lee2022preservation} & 69.11 & 80.43 & 60.65 & 73.20 & 38.13 & 48.03 & 28.95 & 36.31 & 28.39 & 36.41 & 24.67 & 32.16 \\
FedProc~\cite{mu2023fedproc}  & 49.71 & 73.54 & 50.91 & 70.10 & 24.20 & 44.52 & 23.74 & 36.90 & 12.69 & 28.84 & 15.00 & 23.74 \\
FedDecorr~\citep{shi2022towards} & 71.29 & 78.99 & 60.01 & 72.38 & 39.42 & 48.45 & 30.56 & 38.20 & 27.93 & 33.51 & 24.34 & 30.28 \\
\textbf{FedRCL (ours)} & \textbf{75.94} & \textbf{84.67} & \textbf{72.93} & \textbf{81.71} & \textbf{50.83} & \textbf{59.07} & \textbf{37.23} & \textbf{46.98} & \textbf{32.09} & \textbf{40.87} & \textbf{30.44} & \textbf{36.44} \\
\bottomrule
\end{tabular}}
\end{center}
\vspace{-4mm}
\end{table*}

%% file: sections/exp.tex

\section{Experiment}
\label{sec:exp}

\subsection{Experimental setup}

\paragraph{Datasets and baselines}

We employ three standard benchmarks for experiments: CIFAR-10, CIFAR-100~\citep{krizhevsky2009learning}, and Tiny-ImageNet~\citep{le2015tiny}, 
covering various levels of data heterogeneity and participation rates. 
We generate \textit{i.i.d.} datasets by randomly assigning training examples to each client without replacement. 
For non-\textit{i.i.d.} cases, we simulate data heterogeneity by sampling label ratios from a Dirichlet distribution with a symmetric parameter $\alpha \in$ \{0.05, 0.1, 0.3, 0.6\} following \cite{hsu2019measuring}.
The participation ratio is 5\% out of 100 distributed clients unless stated otherwise.
Following existing literature, each client holds an equal number of examples. 
For evaluation, we use the complete test set for each dataset and measure the accuracy achieved at the 500$^{\text{th}}$ and 1,000$^{\text{th}}$ rounds.
We compare our method, dubbed as FedRCL, with several state-of-the-art federated learning techniques, which include FedAvg~\citep{mcmahan2017communication}, FedAvg + FitNet~\citep{romero2014fitnets}, FedProx~\citep{li2020federated}, MOON~\citep{li2021model}, FedMLB~\citep{kim2022multi}, FedLC~\citep{zhang2022Federated}, FedNTD~\citep{lee2022preservation},  FedProc~\citep{mu2023fedproc}, and FedDecorr~\citep{shi2022towards}.



\vspace{-2mm}
\paragraph{Implementation details}

We adopt a ResNet-18 as the backbone network, where we replace the batch normalization with the group normalization~\citep{wu2018group} as suggested in~\cite{hsieh2020non}.
We trained the model from scratch, using the SGD optimizer with a learning rate of 0.1, an exponential decay parameter of 0.998, a weight decay of 0.001, and no momentum, following prior works~\cite{acar2021federated, xu2021fedcm, kim2022multi}.
The number of local training epochs is set to 5 and the batch size is adjusted to ensure a total of 10 local iterations at each local epoch throughout all experiments.
We apply contrastive learning to conv1, conv2\_x, conv3\_x, conv4\_x, and conv5\_x layers.
Other hyperparameter settings are as follows for all experiments unless specified otherwise: $\lambda = 0.7$, $\beta = 1$, and $\tau = 0.05$.
We used the PyTorch framework~\citep{paszke2019pytorch} for implementation and executed on NVIDIA A5000 GPUs.
Please refer to the supplementary document for further details about our implementation.

\vspace{2mm}
\subsection{Results}

We compare the proposed method, FedRCL, with numerous client-side federated learning baselines~\citep{mcmahan2017communication, romero2014fitnets, li2020federated, li2021model, kim2022multi, lee2022preservation, mu2023fedproc, shi2022towards, zhang2022Federated} on the CIFAR and Tiny-ImageNet datasets.
Table~\ref{tab:cifar100} demonstrates that our framework outperforms all other existing algorithms by large margins on all datasets and experiment settings.
Among the baselines, MOON presents meaningful performance improvement, but its gains are marginal under severe data heterogeneity, \eg, $\alpha=0.05$.
FedLC employs adaptive label margin to mitigate local deviations, but its impact on performance is limited.
FedDecorr exhibits minor improvement but degraded performance in some settings.
This implies that the blind mitigation of dimensional collapse is not necessarily helpful for FL.
Compared to other works, our algorithm achieves significant performance improvements in all datasets, regardless of the level of data heterogeneity.


\subsection{Analysis}
\label{sec:analysis}


\begin{table}[t!]
\begin{center}
\caption{
Ablative results of contrastive training in the non-\textit{i.i.d.} settings on the CIFAR-100 dataset.
}
\label{tab:cl_abl}
\vspace{-2mm}
\hspace{-0.2cm}
\setlength{\tabcolsep}{1.7mm}
\scalebox{0.85}{
\begin{tabular}{lcccccc}
\toprule
& \multicolumn{2}{c}{$\alpha = 0.05$} & \multicolumn{2}{c}{$\alpha = 0.1$}  & \multicolumn{2}{c}{$\alpha = 0.3$} \\
& {500R} & {1000R} & {500R} & {1000R} & {500R} & {1000R}   \\
\midrule
Baseline & 31.20 & 39.86 & 36.65 & 43.04 & 41.70 & 47.47 \\
FedSCL & 21.22 & 42.93 & 30.93 & 48.09 & 41.54 & 51.70 \\
FedCL & 35.29 & 41.45 & 40.39 & 45.91 & 45.99 & 50.16 \\
\textbf{FedRCL (ours)}  & \textbf{43.71} & \textbf{54.63} & \textbf{49.82} & \textbf{58.23} & \textbf{57.89} & \textbf{63.46} \\
\bottomrule
\end{tabular}
}
\end{center}
\vspace{-6mm}
\end{table}


\paragraph{Low participation rate and large-scale clients}
We validate our framework in more challenging scenarios with lower client participation rates and a larger number of distributed clients.
Table~\ref{tab:large_scale} presents the robust performance improvement of FedRCL on three benchmarks, where a participation rate is 0.02 and the number of clients is set to one of $\{100, 500\}$.
All methods suffer from performance degradation, compared with the results in Table~\ref{tab:cifar100}, due to the reduced client data, increased data disparity, and lower participation rate. 
Particularly, FedProc experiences a significant performance drop, because it relies on the global class prototypes aggregated from participating clients at each round, which may not be accurate in extremely low participation settings.
MOON exhibits performance degradation compared to FedAvg in some challenging configurations, partly because it utilizes outdated previous local models due to the sparse participation of local clients.
Despite these challenges, FedRCL consistently demonstrates promising performance on all the tested datasets.

\vspace{-2mm}
\paragraph{Contrastive learning strategies}
Table~\ref{tab:cl_abl} compares the effectiveness of various contrastive learning strategies, all employing multi-level contrastive training for fair comparisons.
FedSCL is an ablative model of our framework, which incorporates a na\"ive supervised contrastive loss into the FedAvg baseline.
While FedSCL improves upon the baseline in general, its gains are moderate and even negative in the early stage of training.
In contrast, our full framework consistently enhances performance throughout the entire learning process, as also observed in Figure~\ref{fig:teaser}.
We also employ another variant, denoted as FedCL, which adopts a self-supervised contrastive loss~\cite{InfoNCE}, but its benefits are not salient.
This is partly because the objective in \cite{InfoNCE} does not directly align with the reduction of local deviations.


\begin{table}[t!]
\begin{center}
\caption{
Ablative results of multi-level contrastive training in various non-\textit{i.i.d.} settings on the CIFAR-100 dataset.
}
\label{tab:layer_abl}
\vspace{-2mm}
\setlength{\tabcolsep}{2.3mm}
\scalebox{0.85}{
\begin{tabular}{lcccc}
\toprule
& \multicolumn{1}{c}{$\alpha = 0.05$} & \multicolumn{1}{c}{$\alpha = 0.1$}  & \multicolumn{1}{c}{$\alpha = 0.3$} & \multicolumn{1}{c}{$\alpha = 0.6$}  \\
\midrule
Baseline & 39.86 & 43.04 & 47.47 & 49.29 \\
Last-layer only & 46.36 & 52.94 & 57.09 & 58.20 \\
Multi-layers (ours) & \textbf{54.63} & \textbf{58.23} & \textbf{63.46} & \textbf{64.06} \\
\bottomrule
\end{tabular}
}
\end{center}
\vspace{-3mm}
\end{table}

\vspace{-2mm}

\paragraph{Multi-level contrastive learning}
Table~\ref{tab:layer_abl} presents the ablative results of FedRCL on CIFAR-100, where the proposed contrastive learning is applied only to the last-layer feature outputs.
The results show that FedRCL benefits from the contrastive learning on intermediate representations.


%


\begin{figure}[t!]
        \centering
        \includegraphics[width=0.9\linewidth]{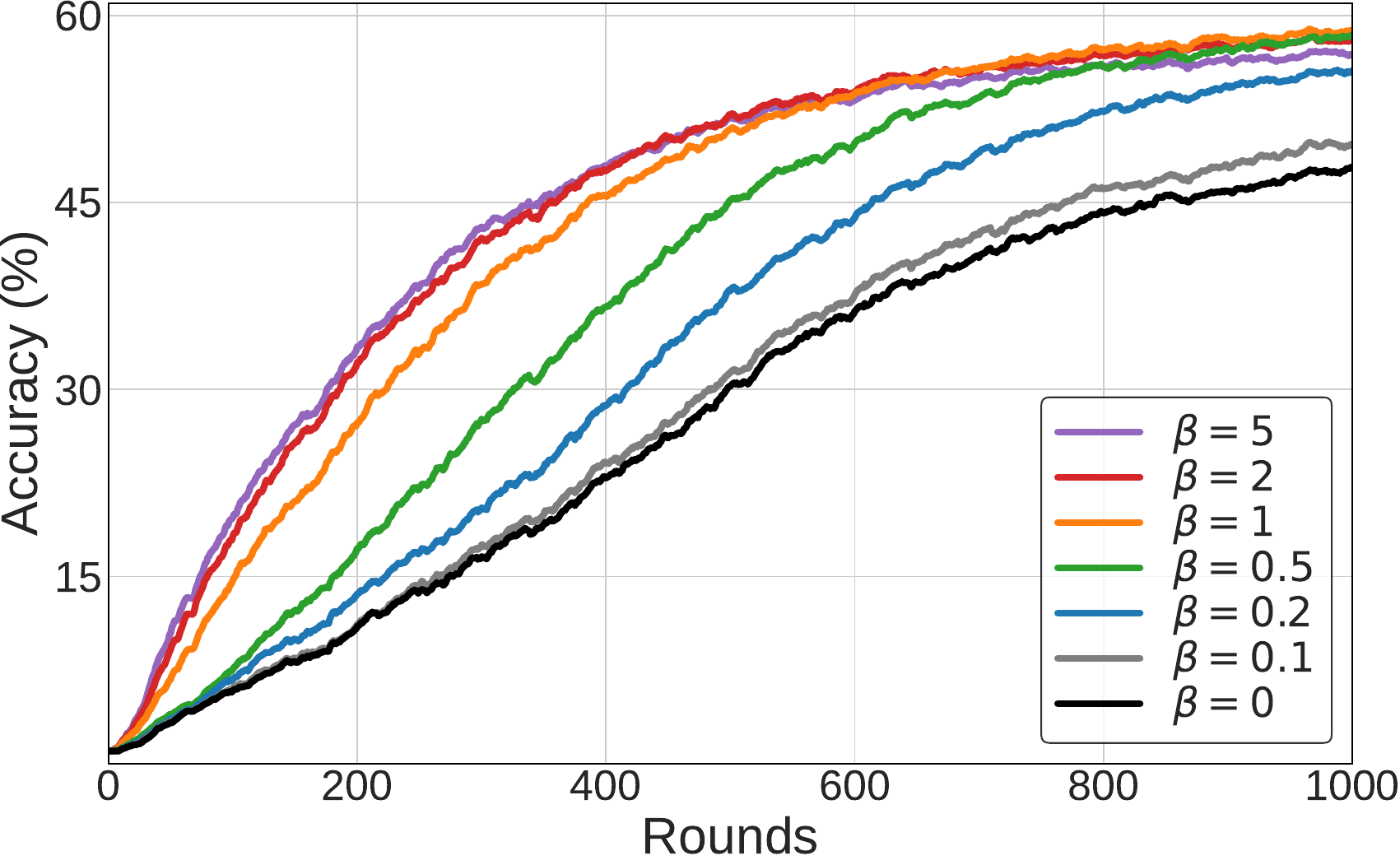}
    \vspace{-0mm}
    \caption{Ablative results by varying the weight of the divergence penalty ($\beta$), which exhibit stability across its wide range.
    }
    \label{fig:rcl_beta_cifar}
    \vspace{-0mm}
\end{figure}

\vspace{-2mm}
\paragraph{Sensitivity of the divergence penalty}
We study the impact of $\beta$ in Eq.~(\ref{eq:rcl}) on the performance of FedRCL under a non-\textit{i.i.d.} setting with $\alpha=0.1$.
Figure~\ref{fig:rcl_beta_cifar} illustrates that a large $\beta$ leads to early convergence and the improvements are consistent over a wide range of $\beta$, although its excessively high values marginally degrades the final performances.


\begin{table}[t!]
\begin{center}
\setlength{\tabcolsep}{1.5mm}
\caption{Integration of FedRCL into various server-side federated learning approaches under a non-\textit{i.i.d.} setting ($\alpha=0.3$).
\label{tab:integration_server}
\vspace{-2mm}
}
\scalebox{0.85}{
\begin{tabular}{lccc}
\toprule
Method &  \multicolumn{1}{c}{CIFAR-10}  &\multicolumn{1}{c}{CIFAR-100}  &\multicolumn{1}{c}{Tiny-ImageNet}  \\ 
\midrule
FedAvgM~\cite{hsu2020federated}     & 85.48 &  53.29  & 38.51\\
FedAvgM + {FedRCL}     & \textbf{88.51}  &  \textbf{64.61}  & \textbf{47.23}  \\                   
\midrule
FedADAM~\cite{reddi2021adaptive} & 81.82  &  52.81  & 39.74 \\
FedADAM + {FedRCL}    & \textbf{85.69}  &  \textbf{57.84}  & \textbf{41.57}  \\                     
\midrule
FedACG~\cite{kim2022communicationefficient}       & 89.10  & 62.51  & 46.31 \\
FedACG + {FedRCL}    & \textbf{89.67} &  \textbf{66.38}  & \textbf{47.97} \\                    
\bottomrule
\end{tabular}}
\end{center}
\vspace{-2mm}
\end{table}

\vspace{-2mm}
\paragraph{Combination with server-side optimization methods}
Our approach is orthogonal to server-side algorithms, which allows seamless combinations of FedRCL and the server-side techniques such as FedAvgM~\citep{hsu2019measuring}, FedADAM~\citep{reddi2021adaptive}, and FedACG~\citep{kim2022communicationefficient}.
Table~\ref{tab:integration_server} presents the consistent and promising performance gains by the combinations,

%% file: sections/conclusion.tex
\section{Conclusion}
\label{sec:conclusion}



We presented a novel federated learning approach to address the challenges of data heterogeneity effectively.
We initiated our investigation by analyzing gradient deviations at each local model and showed that the SCL objective mitigates the local deviations, but it entails representation collapses and limited transferability.
To tackle this issue, we proposed a federated relaxed contrastive learning framework that successfully prevents representation collapses, which is further enhanced by encompassing all levels of intermediate feature representations.
We demonstrated the superiority and robustness of our framework through extensive experiments and analyses.


\paragraph{Acknowledgement}
This work was supported by the National Research Foundation of Korea grants [No. 2022R1A2C3012210; No. 2022R1A5A7083908; No. 2021M3A9E4080782] and the IITP grants [2021-0-02068; 2021-0-01343] funded by the Korea government (MSIT).

%% file: sections/supple.tex

\setcounter{section}{0}
\setcounter{table}{0}
\setcounter{figure}{0}
\setcounter{definition}{0}
\setcounter{proposition}{0}

\renewcommand\thesection{\Alph{section}}
\renewcommand\thetable{\Alph{table}}
\renewcommand\thefigure{\Alph{figure}}

\newpage
\onecolumn
\section{Proof of Proposition~\ref{prop:local_deviation}}
\label{sec:proof_prop1}
\begin{definition}[\textbf{Sample-wise deviation bound}] \label{def:deviation_bound}
Let  $\mathbf{x} \in \mathcal{O}_r$  denote a training example belonging to class $r$. The sample-wise deviation bound is given by
\begin{align}
\label{eq:deviation_bound_supple}
D(\mathbf{x}) 
& = \frac{\left(1-{P_{r}^{(r)}}\right) {\Phi_{r} |\mathcal{O}_r| S_r(\mathbf{x})}}{\sum\limits_{j \neq r} {P}_{r}^{(j)} {{\Phi_{j}}} |\mathcal{O}_j| S_j(\mathbf{x})},
\end{align}
%
where $P_{z}^{(y)}=\frac{1}{|\mathcal{O}_y|}\sum_{i\in \mathcal{O}_y} p_{z}(\mathbf{x}_{i})$ is the average prediction score of the samples in a class $y$ to a class $z$, ${\Phi_{y}} = \frac{1}{|\mathcal{O}_y|}\sum_{i \in \mathcal{O}_y}{\left\|\phi{(\mathbf{x}_i)}\right\|_{2}}$ is the average feature norm of the examples in class $y$, and $S_y(\mathbf{x}) = \frac{1}{|\mathcal{O}_y|}\sum_{i \in \mathcal{O}_y}{\langle \phi{(\mathbf{x})},\phi{(\mathbf{x}_i)} \rangle}$ denotes the average feature similarity to a sample $\mathbf{x}$. 
\end{definition}

\begin{proposition} 
If $D(\mathbf{x}) \ll 1$, the local updates of the parameters in classification layer, $\{\Delta \psi_{y}\}_{y \in \mathcal{Y}}$, are prone to deviate from the desirable direction, \ie, $\Delta \psi_{r}\phi(\mathbf{x}) < 0$ and $\exists j \neq r$ such that $\Delta \psi_{j}\phi(\mathbf{x}) > 0$ for $\mathbf{x} \in \mathcal{O}_r$.
\label{prop:local_deviation}
\end{proposition}
To minimize the classification error for $\mathbf{x} \in \mathcal{O}_r$, we expect $ \Delta \psi_{r} \phi(\mathbf{x}) > 0$ and $\Delta \psi_{j} \phi(\mathbf{x}) < 0$ for all $j \neq r$, increasing the probability $p_{r}(\mathbf{x})=\frac{\exp{(\psi_{r} \phi(\mathbf{x}))}}{\sum_{k \in \mathcal{Y}}\exp(\psi_{k} \phi(\mathbf{x}))}$.
Following~\cite{zhang2022Federated}, we derive the update process for $\psi$ by

\begin{align}
\Delta \psi_{r} & = \eta\sum_{\mathbf{x}_i \in \mathcal{O}_r}
\left(1-{p_{r}(\mathbf{x}_i)}\right){\phi{(\mathbf{x}_i)}}-\eta \sum_{j \neq r} \sum_{\mathbf{x}_i \in \mathcal{O}_j}{p_{r}(\mathbf{x}_i}) {\phi{(\mathbf{x}_i)}} \nonumber\\
& \approx \eta\left(1-P_{r}^{(r)}\right)\sum_{\mathbf{x}_i \in \mathcal{O}_r}{\phi{(\mathbf{x}_i)}}-\eta \sum_{j \neq r} P_{r}^{(j)} \sum_{\mathbf{x}_i \in \mathcal{O}_j}{\phi{(\mathbf{x}_i)}},
\end{align}
where $\eta$ is a learning rate.
Then, $\Delta \psi_{r} \phi(\mathbf{x})$ can be formulated as
%
\begin{align}
\label{eq:update_weight}
    &\Delta \psi_{r} \phi(\mathbf{x}) 
    = \eta\left(1-P_{r}^{(r)}\right)\sum_{\mathbf{x}_i \in \mathcal{O}_r}{\phi{(\mathbf{x}_i)} \cdot \phi(\mathbf{x})}-\eta \sum_{j \neq r} P_{r}^{(j)} \sum_{\mathbf{x}_i \in \mathcal{O}_j}{\phi{(\mathbf{x}_i)} \cdot \phi(\mathbf{x})} \nonumber\\
    &=\eta\left(1-P_{r}^{(r)}\right)\frac{\left\|\phi{(\mathbf{x})}\right\|_{2}}{|\mathcal{O}_r|}\sum_{\mathbf{x}_i \in \mathcal{O}_r}{\left\|\phi{(\mathbf{x}_i)}\right\|_{2}} \sum_{\mathbf{x}_j \in \mathcal{O}_r}{\langle\phi{(\mathbf{x}_j)} ,\phi(\mathbf{x})\rangle}-\eta \sum_{j \neq r} P_{r}^{(j)} \frac{\left\|\phi{(\mathbf{x})}\right\|_{2}}{|\mathcal{O}_j|}\sum_{\mathbf{x}_i \in \mathcal{O}_j}{\left\|\phi{(\mathbf{x}_i)}\right\|_{2}} \sum_{\mathbf{x}_j \in \mathcal{O}_j}{\langle\phi{(\mathbf{x}_j)} ,\phi(\mathbf{x})\rangle}\nonumber\\
    &=\eta\left(1-P_{r}^{(r)}\right)\left\|\phi{(\mathbf{x})}\right\|_{2}{\Phi_{r}} {|\mathcal{O}_r|} S_r(\mathbf{x})-\eta \sum_{j \neq r} P_{r}^{(j)} \left\|\phi{(\mathbf{x})}\right\|_{2}{\Phi_{j}}{|\mathcal{O}_j|} S_j(\mathbf{x})\nonumber \\
    &=\eta \left\|\phi{(\mathbf{x})}\right\|_{2} \Bigg(\sum_{j \neq r} P_{r}^{(j)} {\Phi_{j}}{|\mathcal{O}_j|} S_j(\mathbf{x})\Bigg)
    \Bigg( \underbrace{\frac{\left(1-{P_{r}^{(r)}}\right) {\Phi_{r} |\mathcal{O}_r| S_r(\mathbf{x})}}{\sum\limits_{j \neq r} {P}_{r}^{(j)} {{\Phi_{j}}} |\mathcal{O}_j| S_j(\mathbf{x})}}_{D(\mathbf{x})} -1 \Bigg),
\vspace{-2mm}
\end{align}
where the deviation bound $D(\mathbf{x})$ in~\cref{def:deviation_bound} is derived.
For the second equality, we assume that the cosine similarity of different $\phi(\mathbf{x})$ is independent with the $L_2$-norm of $\phi(\mathbf{x})$. 
In this equation, $\Delta \psi_{r} \phi(\mathbf{x})$ becomes negative when $D(\mathbf{x}) \ll 1$,\footnote{Due to the common practice of employing activation functions like ReLU, the feature output $\phi(\mathbf{\cdot})$ is always non-negative, and consequently, the average feature similarity $S_y(\mathbf{\cdot})$ is also non-negative for any $y \in \mathcal{Y}$. 
This indicates that the sign of $\Delta \psi_{r} \phi(\mathbf{x})$ is solely affected by $D(\mathbf{x})$.} which suggests that the local updates are more likely to deviate from the expected direction with a lower value of $D(x)$.
%
%

Similarly, $\Delta \psi_{j} \phi(\mathbf{x})$ is described as
%
\begin{align}
{\Delta \psi_{j} \phi(\mathbf{x})} 
& = { \eta\left\|\phi{(\mathbf{x})}\right\|_{2}}\bigg( {\Phi_{j}}{|\mathcal{O}_j|} S_j(\mathbf{x})- \sum_{k \in \mathcal{Y}} P_{j}^{(k)} {\Phi_{k}}{|\mathcal{O}_k|}S_k(\mathbf{x})\bigg).
\label{eq:update_weight_theothers}
\end{align}

By taking the average of \cref{eq:update_weight_theothers} over all classes excluding the class $r$, we get

\begin{align}
\label{eq:update_weight_j}
\frac{1}{|\mathcal{Y}|-1}\sum_{j \neq r} {\Delta \psi_{j} \phi(\mathbf{x})} 
& = \frac{{\eta \left\|\phi{(\mathbf{x})}\right\|_{2}}}{|\mathcal{Y}|-1} \left(\sum_{j \neq r} {\Phi_{j}}{|\mathcal{O}_j|} S_j(\mathbf{x})-\sum_{j \neq r} \sum_{k \in \mathcal{Y}} P_{j}^{(k)} {\Phi_{k}}{|\mathcal{O}_k|}S_k(\mathbf{x})\right) \nonumber\\
& = \frac{{\eta \left\|\phi{(\mathbf{x})}\right\|_{2}}}{|\mathcal{Y}|-1} \left(\sum_{j \neq r} {\Phi_{j}}{|\mathcal{O}_j|} S_j(\mathbf{x})-\sum_{k \in \mathcal{Y}} \sum_{j \neq r}  P_{j}^{(k)} {\Phi_{k}}{|\mathcal{O}_k|}S_k(\mathbf{x})\right) \nonumber\\
& = \frac{{\eta \left\|\phi{(\mathbf{x})}\right\|_{2}}}{|\mathcal{Y}|-1} \left(\sum_{k \neq r} {\Phi_{j}}{|\mathcal{O}_k|} S_k(\mathbf{x})-\sum_{k \in \mathcal{Y}}   (1 - P_{r}^{(k)}) {\Phi_{k}}{|\mathcal{O}_k|}S_k(\mathbf{x})\right)\nonumber\\
&= \frac{-{\eta \left\|\phi{(\mathbf{x})}\right\|_{2}}}{|\mathcal{Y}|-1} \left( {\Phi_{r}} {|\mathcal{O}_r|}S_r(\mathbf{x})- \sum_{k \in \mathcal{Y}} P_{r}^{(k)} {\Phi_{k}}{|\mathcal{O}_k|}S_k(\mathbf{x})\right) \nonumber\\
&=\frac{-{\eta \left\|\phi{(\mathbf{x})}\right\|_{2}}}{|\mathcal{Y}|-1}{\Bigg(\sum_{j \neq r} P_{r}^{(j)} {\Phi_{j}}{|\mathcal{O}_j|} S_j(\mathbf{x})\Bigg)
\Bigg( \underbrace{\frac{\left(1-{P_{r}^{(r)}}\right) {\Phi_{r} |\mathcal{O}_r| S_r(\mathbf{x})}}{\sum\limits_{j \neq r} {P}_{r}^{(j)} {{\Phi_{j}}} |\mathcal{O}_j| S_j(\mathbf{x})}}_{D(\mathbf{x})} -1 \Bigg)},
\vspace{-2mm}
\end{align}
where the same $D(\mathbf{x})$ is derived, suggesting that there exists $j \in \mathcal{Y} \setminus r$ for which $\Delta \psi_{j} \phi(\mathbf{x})$ becomes positive if $D(\mathbf{x}) \ll 1$.
Both~\cref{eq:update_weight,eq:update_weight_j} present that lower values of $D(\mathbf{x})$ are likely to lead to gradient deviations.
 $\square$

\section{Mitigating Local Gradient Deviations via SCL}
\label{sec:proof_prop2}
By Proposition~\ref{prop:local_deviation}, our objective is improving $D(\mathbf{x})$ to prevent local gradient deviations.
Assuming that $\frac{S_r(\mathbf{x})}{ \sum\limits_{j \neq r} S_j(\mathbf{x})} \geq \frac{1}{|\mathcal{Y}|-1}$, we derive the lower bound of $D(\mathbf{x})$ as 
%
\begin{align}
D(\mathbf{x}) 
& = \frac{\left(1-{P_{r}^{(r)}}\right) {\Phi_{r} |\mathcal{O}_r| S_r(\mathbf{x})}}{\sum\limits_{k \neq r} {P}_{r}^{(k)} {{\Phi_{k}}} |\mathcal{O}_k| S_k(\mathbf{x})} \nonumber \\
&=  \frac{S_r(\mathbf{x})}{ \sum\limits_{k \neq r} \min\limits_{j \neq r}\left\{\frac{{P}_{r}^{(k)}  {{\Phi_{k}}} |\mathcal{O}_k| }{ {P}_{r}^{(j)} {{\Phi_{j}}} |\mathcal{O}_j| }  \right\} S_k(\mathbf{x})} (1-{P_{r}^{(r)}}) {\Phi_{r} |\mathcal{O}_r| } \min\limits_{j \neq r}\left\{\frac{1}{ {P}_{r}^{(j)} {{\Phi_{j}}} |\mathcal{O}_j| }  \right\}\nonumber\\
& \geq   \frac{S_r(\mathbf{x})}{ \sum\limits_{j \neq r} S_j(\mathbf{x})} (1-{P_{r}^{(r)}}) {\Phi_{r} |\mathcal{O}_r| } \min\limits_{j \neq r}\left\{\frac{1}{ {P}_{r}^{(j)} {{\Phi_{j}}} |\mathcal{O}_j| }  \right\}\\
& \geq \frac{\left(1-{P_{r}^{(r)}}\right) {\Phi_{r} |\mathcal{O}_r| }}{|\mathcal{Y}|-1} \min\limits_{j \neq r}\left\{\frac{1}{ {P}_{r}^{(j)} {{\Phi_{j}}} |\mathcal{O}_j| }\right\},
\end{align}
which suggests that encouraging each sample to satisfy $\frac{1}{|\mathcal{Y}|-1} \sum\limits_{j \neq r} S_j(\mathbf{x}) - S_r(\mathbf{x}) \leq 0$ increases the difficulty of encountering $D(\mathbf{x}) \ll\ 1$, thereby alleviating local gradient deviations. 
Thus, we formulate the surrogate objective to minimize
\begin{align}
\max \Big(0, \frac{1}{|\mathcal{Y}|-1} \sum\limits_{j \neq r} S_j(\mathbf{x}) - S_r(\mathbf{x}) \Big).
\label{eq:reformulate_objective}
\end{align}

Using $\max\{a_1, \dots, a_n \} \leq LogSumExp(a_1, \dots, a_n )$, the upper bound of the objective is
%
\begin{align}
& \max \Big(0, \frac{1}{|\mathcal{Y}|-1} \sum\limits_{j \neq r} S_j(\mathbf{x}) - S_r(\mathbf{x}) \Big) \nonumber \\
& \leq  \log\Bigg(\exp(0) + \exp\Big(\sum_{j \neq r} \frac{1}{|\mathcal{Y}|-1} S_j(\mathbf{x}) - S_r(\mathbf{x})\Big) \Bigg) \nonumber\\
& =  \log\left(\exp(-S_r(\mathbf{x}))\bigg(\exp( S_r(\mathbf{x})) +   \exp\Big(\sum_{j \neq r}  \frac{1}{|\mathcal{Y}|-1} S_j(\mathbf{x}) \Big) \bigg) \right) \nonumber\\
&= \log\Big( \exp(-S_r(\mathbf{x})) \Big) + \log\bigg(\exp(S_r(\mathbf{x})) +  \exp \Big(\sum_{j \neq r}  \frac{1}{|\mathcal{Y}|-1} S_j(\mathbf{x}) \Big) \bigg)  \nonumber\\
&= -\log\left( \frac{\exp(  S_r(\mathbf{x}))}{\exp(S_r(\mathbf{x})) +  \exp\big(\sum_{j \neq r}  \frac{1}{|\mathcal{Y}|-1} S_j(\mathbf{x}) \big) } \right) \nonumber \\
%
 & = -\log\left( \frac{\exp(  \frac{1}{|\mathcal{O}_r|}\sum_{\mathbf{x}_i \in \mathcal{O}_r}{\langle \phi{(\mathbf{x})},\phi{(\mathbf{x}_i)} \rangle})}{\exp(\frac{1}{|\mathcal{O}_r|}\sum_{\mathbf{x}_i \in \mathcal{O}_r}{\langle \phi{(\mathbf{x})},\phi{(\mathbf{x}_i)} \rangle}) +  \exp(\sum_{j \neq r} \frac{1}{|\mathcal{Y}|-1}  \frac{1}{|\mathcal{O}_j|}\sum_{\mathbf{x}_i \in O_j}{\langle \phi{(\mathbf{x})},\phi{(\mathbf{x}_i)} \rangle} ) } \right) \nonumber\\
&\leq -\log\left( \frac{\exp(  \frac{1}{|\mathcal{O}_r| - 1}\sum_{\mathbf{x}_i \in \mathcal{O}_r \setminus \mathbf{x}}{\langle \phi{(\mathbf{x})},\phi{(\mathbf{x}_i)} \rangle})}{\exp(\frac{1}{|\mathcal{O}_r| - 1}\sum_{\mathbf{x}_i \in \mathcal{O}_r \setminus \mathbf{x}}{\langle \phi{(\mathbf{x})},\phi{(\mathbf{x}_i)} \rangle}) +  \exp(\sum_{j \neq r} \frac{1}{|\mathcal{Y}|-1}  \frac{1}{|\mathcal{O}_j|}\sum_{\mathbf{x}_i \in O_j}{\langle \phi{(\mathbf{x})},\phi{(\mathbf{x}_i)} \rangle} ) } \right) \nonumber\\
& \leq -\log\left( \frac{\exp(  \frac{1}{|\mathcal{O}_r| -1}\sum_{\mathbf{x}_i \in \mathcal{O}_r \setminus \mathbf{x}}{\langle \phi{(\mathbf{x})},\phi{(\mathbf{x}_i)} \rangle})}{\exp(\frac{1}{|\mathcal{O}_r| -1}\sum_{\mathbf{x}_i \in \mathcal{O}_r \setminus \mathbf{x}}{\langle \phi{(\mathbf{x})},\phi{(\mathbf{x}_i)} \rangle}) + \frac{1}{|\mathcal{Y}|-1} \sum_{j \neq r}\exp(   \frac{1}{|\mathcal{O}_j|}\sum_{\mathbf{x}_i \in O_j}{\langle \phi{(\mathbf{x})},\phi{(\mathbf{x}_i)} \rangle} ) } \right)\nonumber \tag{I$^{\ast}$} \label{eq:suppI}\\
& \leq -\log\left( \frac{\exp(  \frac{1}{|\mathcal{O}_r| -1}\sum_{\mathbf{x}_i \in \mathcal{O}_r \setminus \mathbf{x}}{\langle \phi{(\mathbf{x})},\phi{(\mathbf{x}_i)} \rangle})}{\frac{1}{|\mathcal{O}_r| -1}\sum_{\mathbf{x}_i \in \mathcal{O}_r \setminus \mathbf{x}}\exp({\langle \phi{(\mathbf{x})},\phi{(\mathbf{x}_i)} \rangle}) + \frac{1}{|\mathcal{Y}|-1} \sum_{j \neq r} \frac{1}{|\mathcal{O}_j|}\sum_{\mathbf{x}_i \in O_j}\exp(  {\langle \phi{(\mathbf{x})},\phi{(\mathbf{x}_i)} \rangle} ) } \right)\nonumber \tag{II$^{\ast}$}\label{eq:suppII}\\
& \leq -\log\left( \frac{\exp(  \frac{1}{|\mathcal{O}_r| -1}\sum_{\mathbf{x}_i \in \mathcal{O}_r \setminus \mathbf{x}}{\langle \phi{(\mathbf{x})},\phi{(\mathbf{x}_i)} \rangle})}{\sum_{\mathbf{x}_i \neq \mathbf{x}}\exp({\langle \phi{(\mathbf{x})},\phi{(\mathbf{x}_i)} \rangle})} \right)\nonumber\\
& = \frac{-1}{|\mathcal{O}_r| -1}\sum_{\mathbf{x}_i \in \mathcal{O}_r \setminus \mathbf{x}}\log\left( \frac{\exp( {\langle \phi{(\mathbf{x})},\phi{(\mathbf{x}_i)} \rangle})}{\sum_{\mathbf{x}_k \neq \mathbf{x}}\exp({\langle \phi{(\mathbf{x})},\phi{(\mathbf{x}_k)} \rangle})} \right),
\label{eq:scl_form_}
\end{align}
where (\ref{eq:suppI}) and (\ref{eq:suppII}) come from Jensen's inequality.
$\square$

\section{Additional Experiments}

\paragraph{Quantity-based data heterogeneity configurations}
Beside distribution-based data heterogeneity, we additionally employ quantity-based heterogeneity configurations for comprehensive evaluation.
Let assume $M$ training samples are distributed among $N$ clients.
We initially organize the data by class labels and split it into $\gamma\cdot N$ groups, with each group having $\frac{M}{\gamma\cdot N}$ samples.
Note that there is no overlap in the samples held by different clients in these settings.
Our framework consistently exhibits superior performance as evidenced in Table~\ref{tab:quantity-based}, which verifies the robustness of our framework across diverse data heterogeneity scenarios.

\begin{table*}[t!]
\begin{center}
\caption{Results from quantity-based data heterogeneity configurations over 100 distributed clients on the three benchmarks.}
\label{tab:quantity-based}
\setlength\tabcolsep{7.7pt}
\scalebox{0.85}{\begin{tabular}{lcccc|cccc|cccc} 
\toprule
& \multicolumn{4}{c}{CIFAR-10}  & \multicolumn{4}{c}{CIFAR-100}  & \multicolumn{4}{c}{Tiny-ImageNet} \\ 
 & \multicolumn{2}{c}{$\gamma=2$} & \multicolumn{2}{c|}{$\gamma=5$} & \multicolumn{2}{c}{$\gamma=20$} & \multicolumn{2}{c|}{$\gamma=50$} & \multicolumn{2}{c}{$\gamma=20$} & \multicolumn{2}{c}{$\gamma=50$}  \\
Method & {500R} & {1000R} & {500R} & {1000R} & {500R} & {1000R} & {500R} & {1000R} & {500R} & {1000R} & {500R} & {1000R}  \\
\midrule
FedAvg~\cite{mcmahan2017communication} & 37.22 & 52.88 & 71.57 & 82.04 & 37.94 & 44.39 & 44.31 & 50.02& 23.59 & 28.30 & 30.32 & 32.83 \\
FedLC~\citep{zhang2022Federated} & 28.24 & 35.69 & 77.06 & 83.65 & 41.35 & 46.62 & 44.11 & 48.65& \textbf{27.90} & 29.21 & 33.24 & 34.92 \\
FedDecorr~\citep{shi2022towards} & 42.93 & 60.63 & 74.49 & 82.15& 39.63 & 46.40 & 44.62 & 50.30& 22.74 & 27.20 & 29.92 & 32.62\\
\textbf{FedRCL (ours)}  & \textbf{55.01} & \textbf{71.66} & \textbf{81.19} & \textbf{87.66} & \textbf{51.09} & \textbf{59.78} & \textbf{58.05} & \textbf{63.50} & {26.53} & \textbf{33.43} & \textbf{34.18} &\textbf{41.49}\\
\bottomrule
\end{tabular}}
\end{center}
\end{table*}


\paragraph{Integration into server-side optimization approaches}
To supplement Table~\ref{tab:integration_server} in the main paper, we evaluate other recent client-side approaches~\cite{shi2022towards,zhang2022Federated} combined with various server-side algorithms~\cite{hsu2019measuring,reddi2021adaptive,kim2022communicationefficient} for additional comparisons.
As shown in Table~\ref{tab:integration_server_sup}, our framework consistently outperforms FedLC and FedDecorr on top of existing server-side frameworks.

\begin{table}[t]
\begin{center}
\setlength{\tabcolsep}{5mm}
\caption{Integration of client-side approaches into various server-side approaches under non-\textit{i.i.d.} setting ($\alpha=0.3$).
\label{tab:integration_server_sup}
}
\scalebox{0.85}{
\begin{tabular}{lcc|cc|cc}
\toprule
 &  \multicolumn{2}{c}{CIFAR-10}  &\multicolumn{2}{c}{CIFAR-100}  &\multicolumn{2}{c}{Tiny-ImageNet}  \\ 
Method & {500R} & {1000R} & {500R} & {1000R} & {500R} & {1000R}\\
\midrule
FedAvgM~\cite{hsu2020federated} &80.56    & 85.48 &46.98  &  53.29 &36.32 & 38.51\\
FedAvgM + {FedLC}    & 82.03 & 86.41 & 46.96 & 52.91 & 37.76 & 40.50  \\  
FedAvgM + {FedDecorr}    & 80.57 & 85.51 & 46.31 & 53.11 & 34.66 & 36.95 \\  
FedAvgM + \textbf{FedRCL (ours)}    & \textbf{84.62} & \textbf{88.51} & \textbf{60.55} & \textbf{64.61} & \textbf{43.11} & \textbf{47.23} \\                   
\midrule
FedADAM~\cite{reddi2021adaptive} & 75.91 & 81.82 & 47.99 & 52.81  &36.33  & 39.74 \\
FedADAM + {FedLC}     & 77.96 & 82.11 & 49.76 & 53.15 & \textbf{39.04} & 42.12  \\  
FedADAM + {FedDecorr}     & 76.44 & 82.21 & 48.62 & 53.48  & 35.92 & 39.38\\  
FedADAM + \textbf{FedRCL (ours)}  &\textbf{80.71}   & \textbf{85.69}&\textbf{52.86}  &  \textbf{57.84}  & 38.34& \textbf{42.27} \\                     
\midrule
FedACG~\cite{kim2022communicationefficient}       & 85.13 & 89.10 & 55.79 & 62.51 & 42.26 & 46.31 \\
FedACG + {FedLC}     & 85.89 & 89.61 & 57.18 & 62.09 & 43.43 & 44.57 \\  
FedACG + {FedDecorr}     & 85.20 & 89.48 & 57.95 & 63.02 & 43.09 & 44.52 \\  
FedACG + \textbf{FedRCL (ours)} &\textbf{86.43}   & \textbf{89.67} &\textbf{62.82} &  \textbf{66.38}&\textbf{45.97}  & \textbf{47.97} \\                    
\bottomrule
\end{tabular}}
\end{center}
\end{table}


\paragraph{Other backbone networks}

We evaluate FedRCL using different backbone architectures, including VGG-9~\cite{simonyan2014very}, MobileNet-V2~\cite{Sandler2018CVPR}, ShuffleNet~\cite{Zhang2018CVPR}, and SqueezeNet~\cite{iandola2016squeezenet} on CIFAR-100, where we set $\beta$ to 2 for MobileNet and 1 for others.
According to Table~\ref{tab:other_backbone}, FedRCL outperforms other algorithms by large margins regardless of backbone architectures, which shows the generality of our approach.

\begin{table}[t!]
\begin{center}
\caption{
Experimental results with different backbone architecture on the CIFAR-100 dataset under non-\textit{i.i.d.} setting ($\alpha=0.3$).
}
\label{tab:other_backbone}
\setlength{\tabcolsep}{8pt}
\scalebox{0.85}{
\begin{tabular}{lccccc}
\toprule
& SqueezeNet & ShuffleNet & VGG-9 & MobileNet-V2 \\
\midrule
FedAvg~\cite{mcmahan2017communication} & 39.62 & 35.37 & 45.60 & 43.57 \\
+ FitNet~\cite{romero2014fitnets} & 37.78 & 36.18 & 45.35 & 43.89 \\
FedProx~\cite{li2020federated} & 38.86 & 35.37 & 45.32 & 43.09 \\
MOON~\cite{li2021model} & 24.16 & 34.17 & 52.13 & 34.05 \\
FedMLB~\cite{kim2022multi} & 41.95 & 41.61 & 54.36 & 47.09 \\
FedLC~\citep{zhang2022Federated} & 42.35 & 37.79 & 48.46 & 45.51 \\
FedNTD~\cite{lee2022preservation} & 40.33 & 40.14 & 50.78 & 44.85 \\
FedProc~\cite{mu2023fedproc}  & 31.45 & 35.23 & 43.14 & 23.60 \\
FedDecorr~\citep{shi2022towards} & 40.23 & 38.77 & 47.32 & 47.31 \\
\textbf{FedRCL (ours)} & \textbf{49.34} & \textbf{44.50} & \textbf{55.53} & \textbf{51.32} \\
\bottomrule
\end{tabular}
}
\end{center}
\vspace{-3mm}
\end{table}


\paragraph{Larger number of local epochs}
To validate the effectiveness in conditions of more severe local deviations, we evaluate our framework by increasing the number of local epochs to $E=10$.
Table~\ref{tab:local_epochs} presents consistent performance enhancements of FedRCL in the presence of more significant local deviations.

\begin{table}[t!]
\begin{center}
\caption{
Experimental results with an increased number of local epochs ($E=10$) under non-\textit{i.i.d.} setting. 
}
\label{tab:local_epochs}
\setlength{\tabcolsep}{3mm}
\scalebox{0.85}{
\begin{tabular}{lcccc|cccc|cccc}
\toprule
& \multicolumn{4}{c|}{CIFAR-10} & \multicolumn{4}{c|}{CIFAR-100}  & \multicolumn{4}{c}{Tiny-ImageNet} \\
& \multicolumn{2}{c}{$\alpha=0.05$} & \multicolumn{2}{c|}{$\alpha=0.3$} & \multicolumn{2}{c}{$\alpha=0.05$} & \multicolumn{2}{c|}{$\alpha=0.3$} & \multicolumn{2}{c}{$\alpha=0.05$} & \multicolumn{2}{c}{$\alpha=0.3$}  \\
& {500R} & {1000R} & {500R} & {1000R} & {500R} & {1000R} & {500R} & {1000R} & {500R} & {1000R} & {500R} & {1000R}   \\
\midrule
Baseline & 56.80 & 68.52 & 77.79 & 83.78 & 34.64 & 42.35 & 41.47 & 47.49 & 22.38 & 23.65 & 32.49 & 34.58 \\
FedLC~\citep{zhang2022Federated} & 60.81 & 69.59 & 79.58 & 84.71 & 36.83 & 43.99 & 42.7 & 48.04 & 25.73 & 27.51 & 33.38 & 35.30  \\
FedDecorr~\citep{shi2022towards} & 58.34 & 68.64 & 80.55 & 84.91 & 34.91 & 41.84 & 42.73 & 49.25 & 21.48 & 22.54 & 30.65 & 33.06 \\
\textbf{FedRCL (ours)} & \textbf{74.02} & \textbf{78.97} & \textbf{86.58} & \textbf{89.40} & \textbf{49.64} & \textbf{55.91} & \textbf{60.58} & \textbf{64.73} & \textbf{31.01} & \textbf{37.70} & \textbf{44.74} & \textbf{48.51} \\
\bottomrule
\end{tabular}
}
\end{center}
\end{table}

\section{Qualitative Results}
\begin{figure*}[h]
\centering
    \begin{minipage}[t]{0.45\linewidth}
        \centering
        \includegraphics[width=\linewidth]{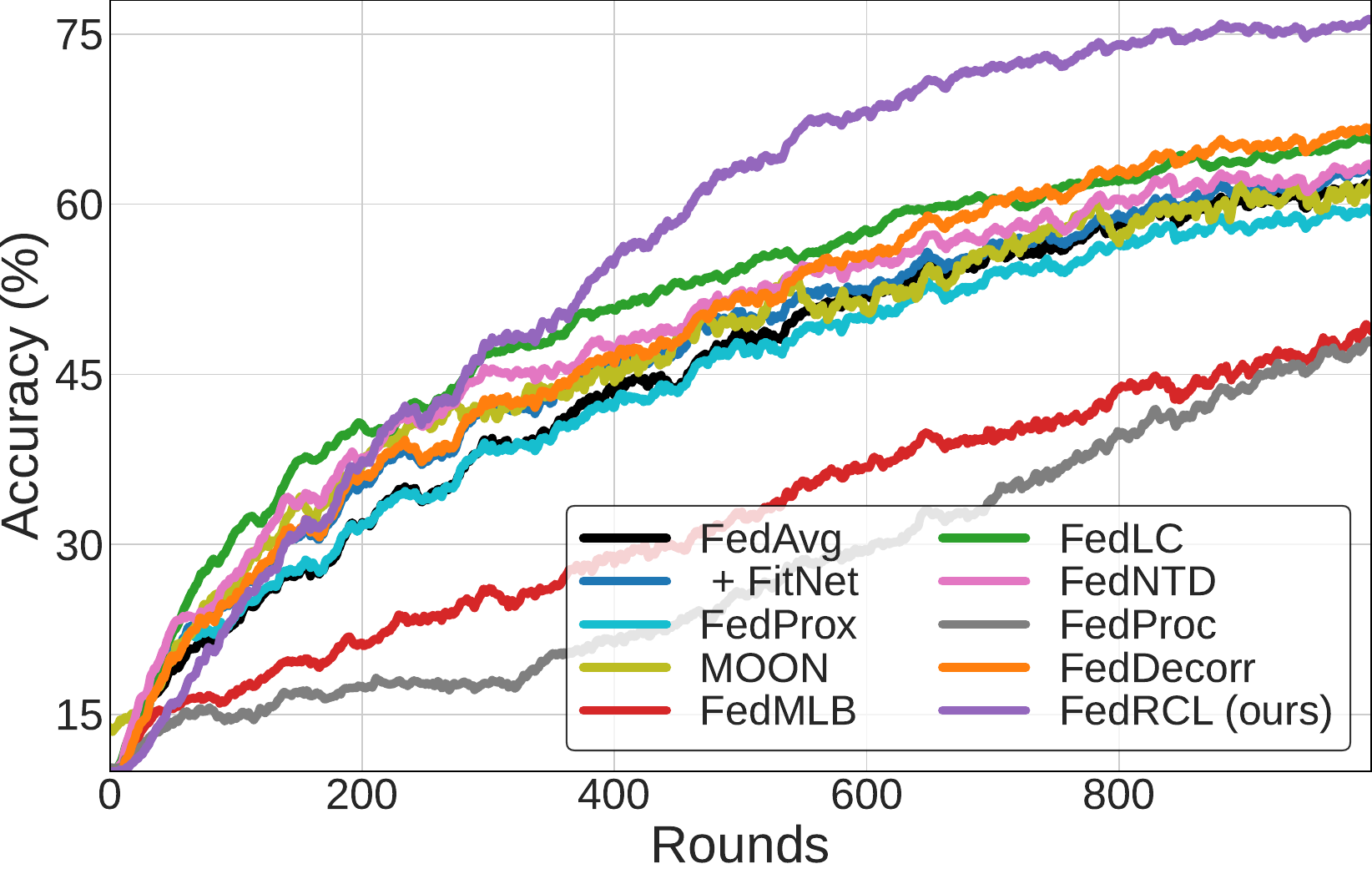}
        \subcaption{CIFAR-10}
        \label{fig:all_methods_cifar100}
    \end{minipage}
    \hspace{5mm}
    \begin{minipage}[t]{0.45\linewidth}
        \centering
        \includegraphics[width=\linewidth]{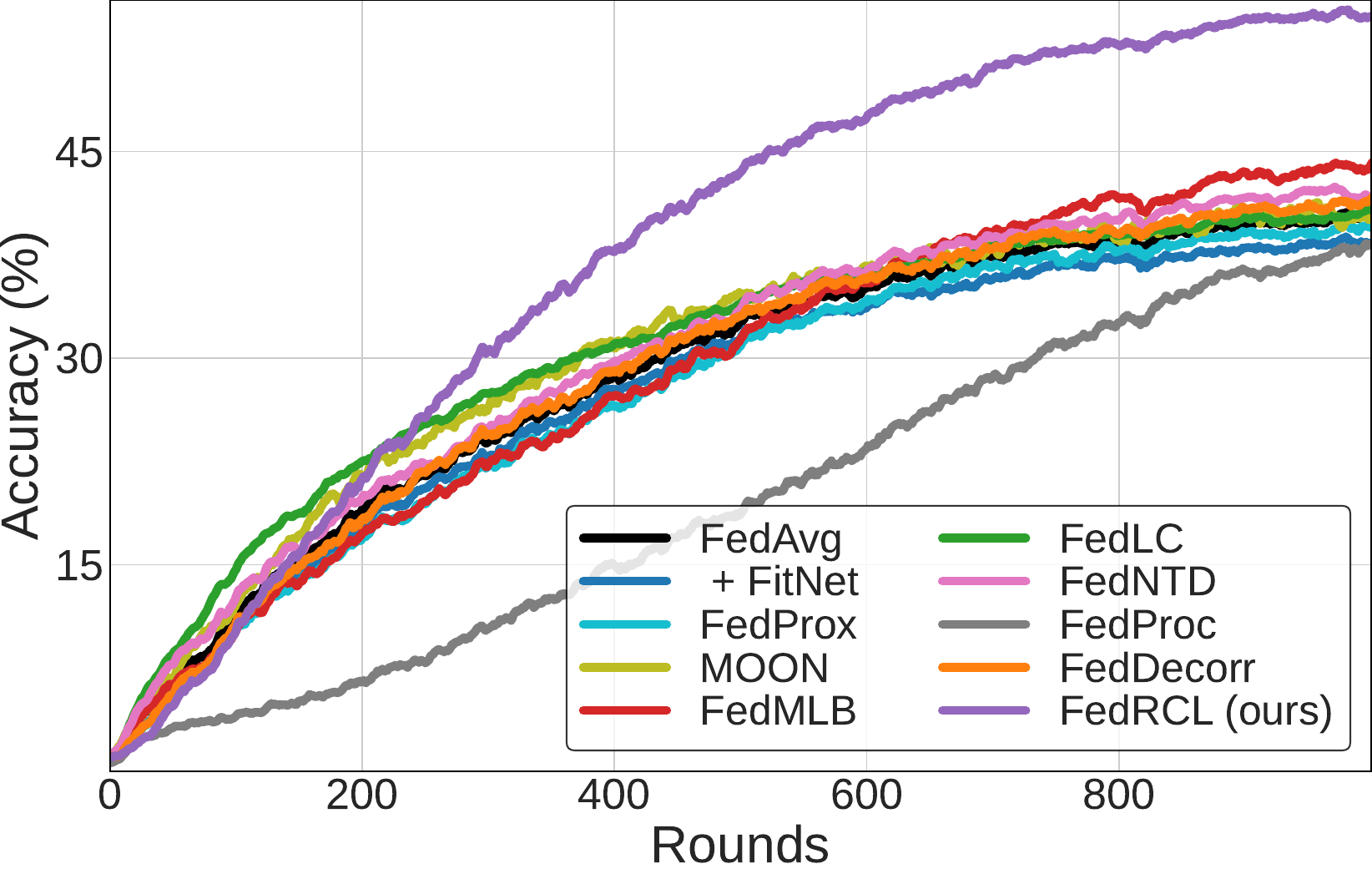}
        \subcaption{CIFAR-100}
        \label{fig:all_methods_tiny}
    \end{minipage}
    \caption{Convergence curve of FedRCL, along with other compared methods, on the CIFAR-10 and CIFAR-100 with non-\textit{i.i.d.} setting ($\alpha=0.05$).
    Accuracy at each round is based on the exponential moving average result with parameter $0.9$.
    }
    \label{fig:all_methods}
\end{figure*}

\paragraph{Convergence plot}
Figure~\ref{fig:all_methods} visualizes the convergence curves of FedRCL and the compared algorithms on CIFAR-10 and CIFAR-100 under non-\textit{i.i.d.} setting ($\alpha=0.05$), where our framework consistently outperforms all other existing federated learning techniques by huge margins throughout most of the learning process.

\begin{figure*}[h]
\centering
    \begin{minipage}[t]{0.45\linewidth}
        \centering
        \includegraphics[width=\linewidth]{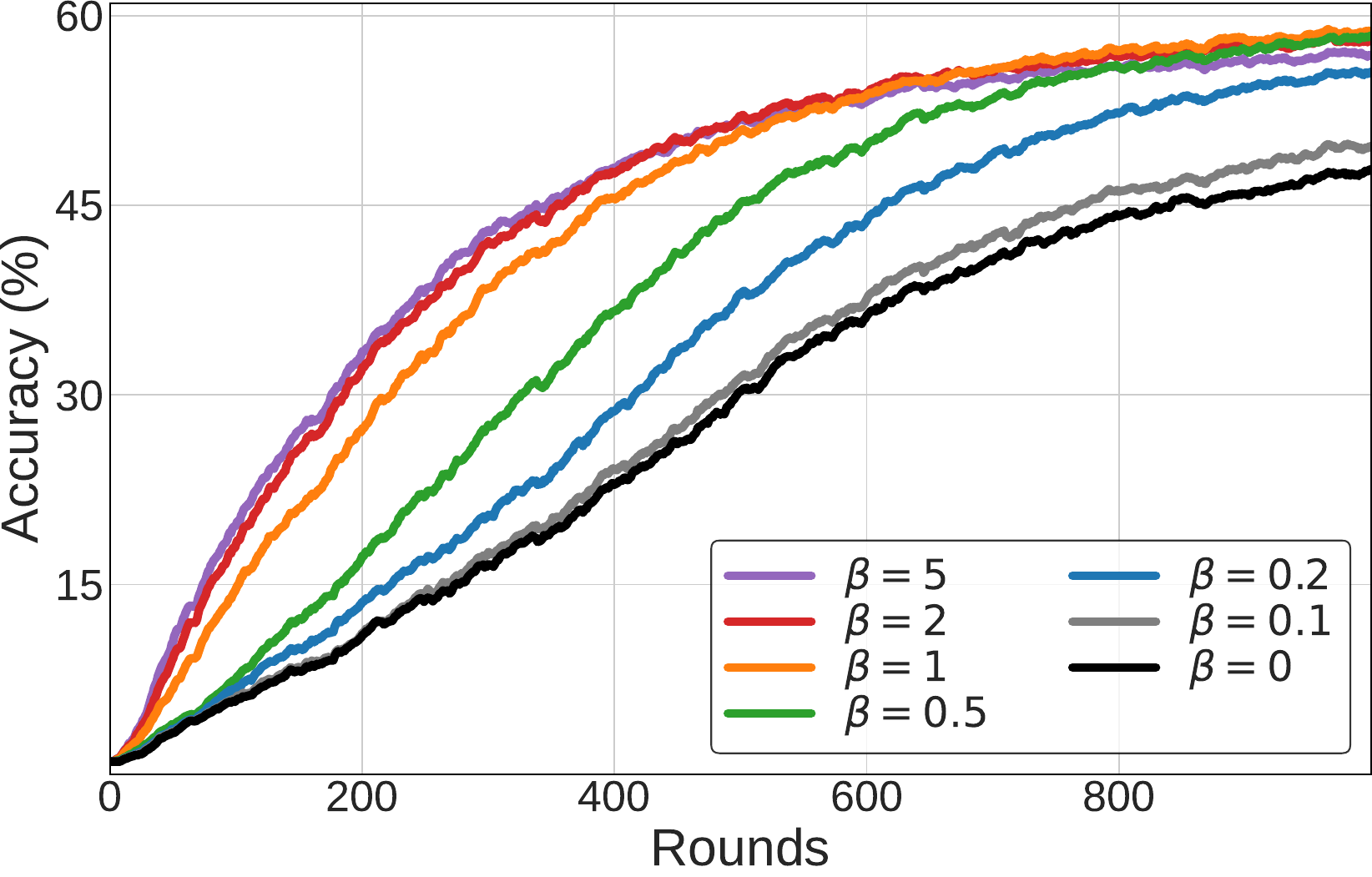}
        \subcaption{$\alpha = 0.1$}
        \label{fig:rcl_beta_cifar_dir01}
    \end{minipage}
    \hspace{5mm}
    \begin{minipage}[t]{0.45\linewidth}
        \centering
        \includegraphics[width=\linewidth]{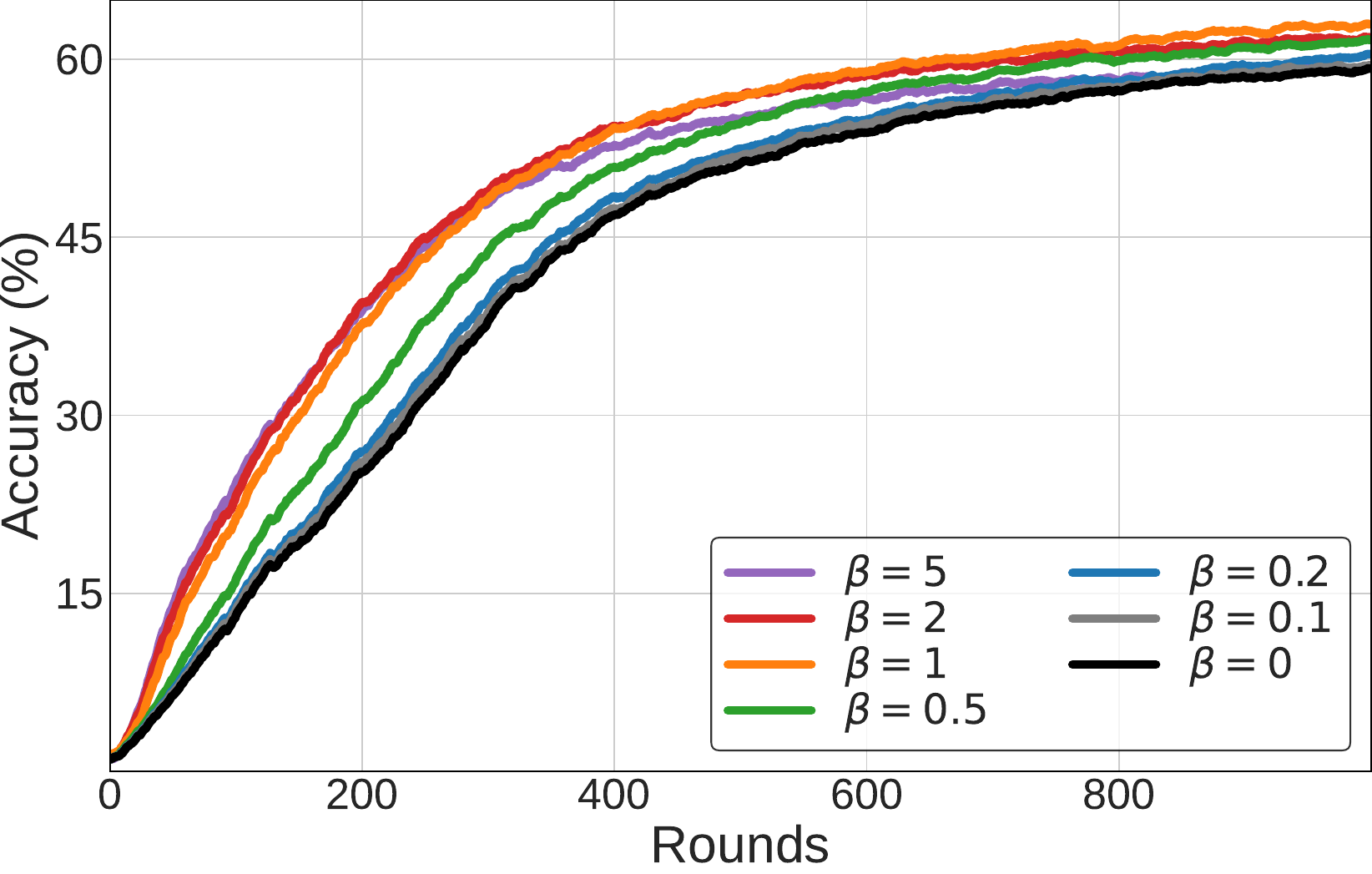}
        \subcaption{$\alpha = 0.3$}
        \label{fig:rcl_beta_cifar_dir03}
    \end{minipage}
    \caption{
    Ablative results by varying the weight of the divergence penalty ($\beta$) on the CIFAR-100 dataset with $\alpha \in \{0.1, 0.3\}$, which exhibits stability across a wide range.
    }
    \label{fig:rcl_beta_cifar_all}
\end{figure*}

\paragraph{Sensitivity on the weight of divergence penalty}
We examine the robustness of our framework by varying the divergence penalty weight $\beta \in \{0, 0.1, 0.2, 0.5, 1, 2, 5\}$ on the CIFAR-100 in non-\textit{i.i.d.} settings.
Figure~\ref{fig:rcl_beta_cifar_all} presents consistent performance enhancements over a wide range of $\beta$, which demonstrates its stability.

\section{Experimental Detail}

\paragraph{Hyperparameter selection}
To reproduce the compared approaches, we primarily follow the settings from their original papers, adjusting the parameters only when it leads to improved performance.
In client-side federated learning approaches, we use 0.001 in FedProx, 0.3 in FedNTD, and 0.01 in FedDecorr, for $\beta$.
We set $\lambda$ to 0.001 in FitNet, while $\lambda_1$ and $\lambda_2$ are both set to 1 in FedMLB.
$\mu$ in MOON and $\tau$ in FedLC are both set to 1.
We adopt $\lambda$ of 0.7, $\beta$ of 1, and $\tau$ of 0.05 in FedRCL.
For server-side algorithms, $\beta$ in FedAvgM is set to 0.4 while $\beta_1$, $\beta_2$, and $\tau$ in FedADAM are set to 0.9, 0.99, and 0.001, respectively.
We use $\lambda$ of 0.85 and $\beta$ of 0.001 in FedACG.

\begin{figure}[h]
    \centering
    \begin{subfigure}{0.23\textwidth}
        \includegraphics[width=\textwidth]{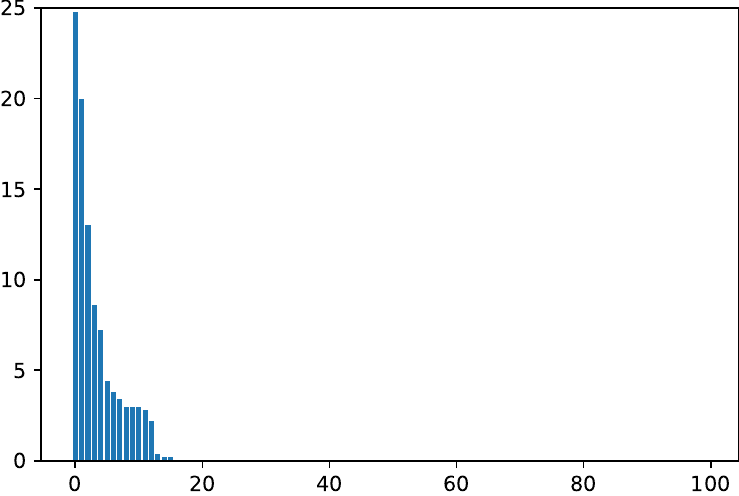}
        \caption{$\alpha = 0.05$}
    \end{subfigure}
    \hfill
    \begin{subfigure}{0.23\textwidth}
        \includegraphics[width=\textwidth]{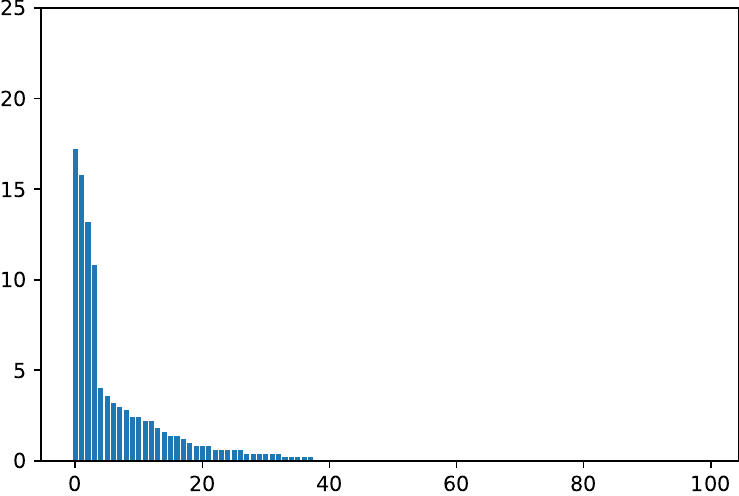}
        \caption{$\alpha = 0.1$}
    \end{subfigure}
    \hfill
    \begin{subfigure}{0.23\textwidth}
        \includegraphics[width=\textwidth]{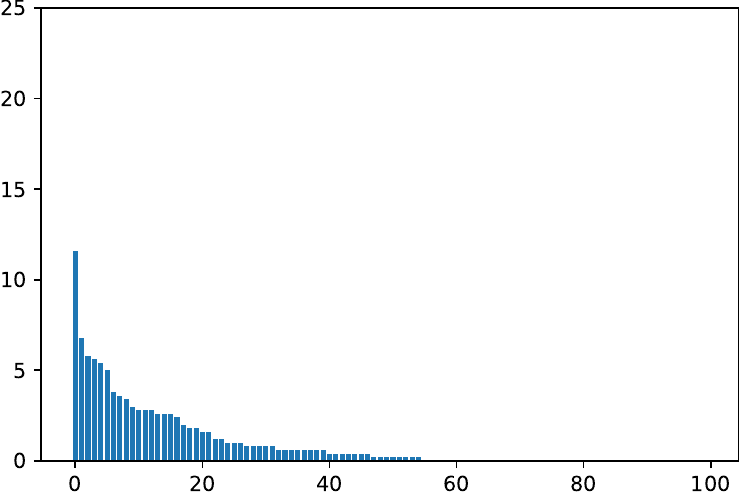}
        \caption{$\alpha = 0.3$}
    \end{subfigure}
    \hfill
    \begin{subfigure}{0.23\textwidth}
        \includegraphics[width=\textwidth]{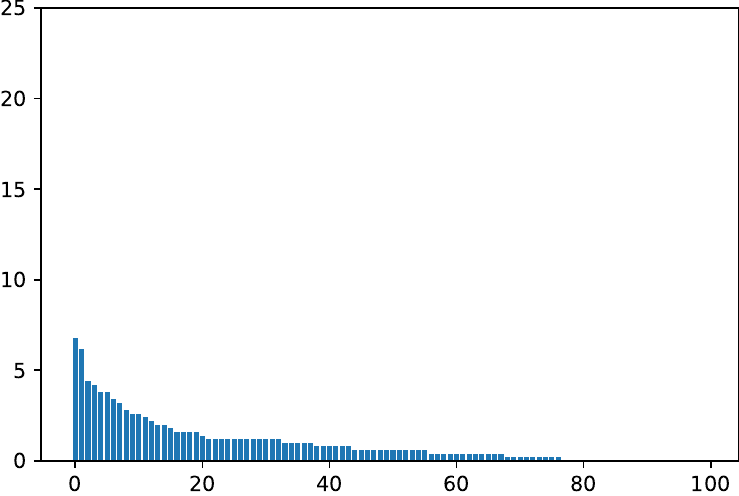}
        \caption{$\alpha = 0.6$}
    \end{subfigure}
    
    \caption{Label distributions at each local client under various heterogeneity configurations with $\alpha \in \{0.05, 0.1, 0.3, 0.6\}$  on the CIFAR-100. 
    $y$-axis represents the ratio of data samples in each class to the total dataset, while $x$-axis is sorted based on the number of samples.
    }
\label{fig:clientdatadist}
\end{figure}

\paragraph{Visualization of local data distribution}
We visualize the local data distribution at each client on the CIFAR-100 under diverse heterogeneity configurations in Figure~\ref{fig:clientdatadist}, where the Dirichlet parameter $\alpha$ is varied by $\{0.05, 0.1, 0.3, 0.6\}$.
Lower values indicate more skewed distributions.

